\documentclass[lettersize,journal]{IEEEtran}
\usepackage{amsmath,amsfonts}
\usepackage{algorithmic}
\usepackage{algorithm}
\usepackage{array}
\usepackage[caption=false,font=normalsize,labelfont=sf,textfont=sf]{subfig}
\usepackage{textcomp}
\usepackage{stfloats}
\usepackage{url}
\usepackage{color}
\usepackage{verbatim}
\usepackage{graphicx}
\usepackage{cite}
\usepackage{amssymb}
\usepackage{mathtools}
\usepackage{amsthm}
\usepackage{multirow}
\usepackage{pifont}

\begin{document}

\title{A Novel Riemannian Sparse Representation Learning Network for Polarimetric SAR Image Classification}

\author{Junfei Shi, ~\IEEEmembership{Member, IEEE}, Mengmeng Nie, Weisi Lin, ~\IEEEmembership{Fellow, IEEE}, Haiyan Jin, ~\IEEEmembership{Member, IEEE}, Junhuai Li, ~\IEEEmembership{Member, IEEE}, Rui Wang, \IEEEmembership{Member, IEEE}
\thanks{Junfei Shi is with  the Department
of Computer Science and Technology, Shaanxi Key Laboratory for Network Computing and Security Technology, Xi'an University of Technology, Xi'an, China.Corresponding author: shijunfei@xaut.edu.cn}
\thanks{Mengmeng Nie, Haiyan Jin and Junhuai Li are with  the Department
of Computer Science and Technology, Shaanxi Key Laboratory for Network Computing and Security Technology, Xi'an University of Technology, Xi'an 710048, China.}
\thanks{Weisi Lin was with the School of Computer Science and Engineering, Nanyang Technological University, Singapore 639798.}
\thanks{Rui Wang is with the School of Artificial Intelligence and Computer Science, Jiangnan University, Wuxi 214122, China. }}

\markboth{Journal of \LaTeX\ Class Files,~Vol.~14, No.~8, May~2024}%
{Shell \MakeLowercase{\textit{et al.}}: A Sample Article Using IEEEtran.cls for IEEE Journals}


\maketitle

\begin{abstract}
Deep learning is an effective end-to-end method for Polarimetric Synthetic Aperture Radar(PolSAR) image classification, but it lacks the guidance of related mathematical principle and is essentially a black-box model. In addition, existing deep models learn features in Euclidean space, where PolSAR complex matrix is commonly converted into a complex-valued vector as the network input, distorting matrix structure and channel relationship. However, the complex covariance matrix is Hermitian positive definite (HPD), and resides on a Riemannian manifold instead of a Euclidean one. Existing methods cannot measure the geometric distance of HPD matrices and easily cause some misclassifications due to inappropriate Euclidean measures. To address these issues, we propose a novel \textbf{R}iemannian \textbf{S}parse Representation \textbf{L}earning Network (SRSR\_CNN) for PolSAR images. Firstly, a superpixel-based Riemannian Sparse Representation (SRSR) model is designed to learn the sparse features with Riemannian metric. Then, the optimization procedure of the SRSR model is inferred and further unfolded into an SRSRnet, which can automatically learn the sparse coefficients and dictionary atoms. Furthermore, to learn contextual high-level features, a CNN-enhanced module is added to improve classification performance. The proposed network is a Sparse Representation (SR) guided deep learning model, which can directly utilize the covariance matrix as the network input, and utilize Riemannian metric to learn geometric structure and sparse features of complex matrices in Riemannian space. Experiments on three real PolSAR datasets demonstrate that the proposed method surpasses state-of-the-art techniques in ensuring accurate edge details and correct region homogeneity for classification.
\end{abstract}


\section{Introduction}
\IEEEPARstart{F}{ull} Polarimetric Synthetic Aperture Radar (PolSAR) systems are capable of transmitting and receiving electromagnetic waves in various polarimetric modes. In contrast to single SAR \cite{9779320}, PolSAR data is commonly represented by a $3\times3$ complex covariance matrix instead of complex data for each resolution unit, providing a wealth of scattering information. Leveraging these advantages, PolSAR images have found widespread applications in agriculture management\cite{9037201}, military surveillance\cite{9930813}, change detection\cite{8421639}, and urban monitoring\cite{LING2023103541}, among other fields. PolSAR image classification, as a critical application, has garnered significant attention from researchers.

For decades, numerous conventional methods have been proposed for PolSAR image classification, including scattering mechanism based approaches, such as Cloude decomposition\cite{RN25}, Freeman decomposition, etc. In addition, some statistical distribution based approaches are proposed, such as Wishart \cite{9883540}, G0\cite{RN108}, Kummer U\cite{RN109} etc. Furthermore, various machine learning methods are employed to enhance classification performance, including MRF\cite{RN113}, SVM\cite{RN114} and Random Forest\cite{LI2020102032}, among others. However, these traditional methods can only utilize low-level hand-crafted features and lack deep semantic information, limiting their classification performance.

Representation learning\cite{8938739,RN118} is an effective mathematical model-based classification method for image classification. These methods operate under the assumption that each pixel can be described as a linear combination of a set of atoms in an over-complete dictionary. Representation learning methods primarily encompass nearest-regularized subspace (NRS) classification\cite{RN124}, collaborative representation classifier (CRC)\cite{RN125}, and sparse representation classification (SRC)\cite{8844815}. However, when applied to PolSAR data, some approaches\cite{RN120} convert the PolSAR covariance matrix into a column vector as the feature, which distorts matrix structure and channel relationship. Some other approaches\cite{6947057,RN128} combine scattering parameters and statistical distribution as multiple features to improve classification performance. These methods are pixel-wise classification techniques with a cost of time consumption. To learn covariance matrix and reduce computation time, some superpixel-based methods have been developed\cite{rs6087158}. The above-mentioned methods can learn a set of sparse representation to provide accurate discriminating information, while they lack high-level semantics for recognizing entire objects, especially in heterogeneous terrain types.

Deep learning methods, as end-to-end models, can automatically learn high-level semantics, which have been widely applied in PolSAR image classification. The primary deep network models include CNN\cite{RN28}, GCN\cite{9234621}, DNN\cite{7273936}, GAN\cite{9132630}, Transformer\cite{9658539}, and others. Among them, CNN is the most commonly used method due to its local convolution and weight sharing. For most CNN variants, the common operation is to convert the complex matrix into a complex-valued or real-valued vector as the network input. For instance, Zhang et al.\cite{RN86} were the first to propose the CV-CNN, which extends the CNN into the complex-valued domain. Zhang et al.\cite{RN135} expanded the 2D-CNN to 3D-CNN to overcome the limitation of inadequate spatial information in 2D convolution. Later, Tan et al.\cite{RN136} combined spatial information learned from 3DCNN with scattering features extracted from the CV-CNN. Shi et al.\cite{RN137} combined the superpixel-based GCN and 3DCNN to obtain both global and local features. In addition, considering rich scattering characteristics of PolSAR data, some other deep learning methods extract the scattering features as the network input. Cui et al.\cite{RN138} introduced a polarimetric multipath convolutional neural network to adaptively learn the polarization rotation angle. Zhang et al.\cite{RN139} utilized the scattering similarity learning module and texture-based attention module to enhance the discriminative ability of features. Periasamy et al.\cite{PERIASAMY2022113144} investigated the correlation between intrinsic scattering features and the conductivity characteristics to enhance classification performance.

Although these deep learning techniques have garnered outstanding results, they still encounter two challenges for PolSAR image classification.

\begin{itemize}
	\item They cannot learn features from the complex matrix directly. Current deep models typically convert the PolSAR covariance matrix into a vector or complex-value data as the network input, and utilize Euclidean metric to learn features. However, PolSAR complex matrix (known as Hermitian positive definite (HPD) matrix) resides in Riemannian manifold rather than Euclidian space. So, existing methods totally distorts the geometric structure of complex matrices and lose channel correlations.
	\item The second issue is the network learning process is unexplainable. Deep learning methods rely on empirical learning with numerous training samples, adjusting network parameters through gradual training and testing. The absence of guiding principles or theories makes network learning essentially a black box, without understanding of how it works and what features it has learned. Therefore, it is necessary to build an interpreted network that can theoretically guide the network's learning process.
\end{itemize}

For the first issue, since PolSAR covariance matrix is endowed in the Riemannian manifold instead of Euclidean space, some metrics\cite{10282134} in Riemannian space can effectively measure the geometric distance of two matrices. For instance, Ref.\cite{7565529} has proved the Riemannian metric can better describe PolSAR data than Euclidean metric. In addition, we propose a Riemannian Nearest regularization subspace method\cite{9963700} for PolSAR image classification. Further, the complex matrix and multi-feature joint learning method \cite{shi2023complex} is designed to enhance classification performance. However, these methods only define the sparse representation model in Riemannian space without deep learning, which limits the capability of learning high-level features.

For the second issue, some model-based approaches are helpful since they have explained mathematical mechanism and easily ensemble domain knowledge. Some model-based deep learning methods\cite{9690515} have been proposed for natural images, such as CODE-Net\cite{9613794}, ISTAnet\cite{RN144}, deep stacked T-net\cite{9919385},etc. However, these methods cannot be directly applied to PolSAR images without considering geometric structure of complex matrices. As far as we are concerned, there are no related model-guided deep learning methods for PolSAR images.

To address these deficiencies, in this paper, we propose a novel Riemannian sparse representation guided deep learning method (SRSR\_CNN) for PolSAR image classification for the first time. The proposed method combines the advantages of theoretical interpretation from the sparse representation (SR) model and the feature learning ability from the CNN model. Specifically, we first develop a novel Superpixel-based Riemannian Sparse Representation model (SRSR) for PolSAR images, which can model the HPD matrix with a manifold metric to better learn the matrices' geometric structure. Then, the SRSR model is converted into a Superpixel-based Riemannian Sparse Representation Network (SRSRNet) backbone, designed to guide the construction of the learned deep network. Finally, the proposed Riemannian sparse representation learning network (SRSR\_CNN) is an end-to-end framework that can learn sparse features of the HPD matrix directly, as well as high-level semantics by combining the proposed SRSRNet and CNN models.
The main contributions of the proposed method can be summarized into three aspects.
\begin{itemize}
	\item A novel SRSR\_CNN network is proposed for PolSAR image classification, which is an SR-guided learning network rather than blind learning, capable of acquiring model-oriented sparse features with mathematical interpretation.
	\item A superpixel-based Riemannian sparse representation (SRSR) model is developed for modeling the PolSAR covariance matrices in Riemannian space. Based on this model, a lightweight SRSRNet structure is put forward by unfolding the optimization procedure of SRSR model to learn sparse features with fewer training samples.
	\item The proposed SRSR\_CNN combines the SRSRNet and CNN model together with covariance matrices as the input, which can directly learn complex covariance matrix-based high-level semantic features. Experiments on three real PolSAR datasets validate the effectiveness of the proposed SRSR\_CNN.
\end{itemize}

The remainder of this article is structured as follows. Section II provides a detailed introduction to the proposed SRSR\_CNN method. Section III presents the experimental results and analyses. Finally, Section IV concludes the paper.

\section{Proposed method}

\begin{figure*}[ht]
	\vskip 0.2in
	\begin{center}
		\centerline{\includegraphics[height=0.31\textheight]{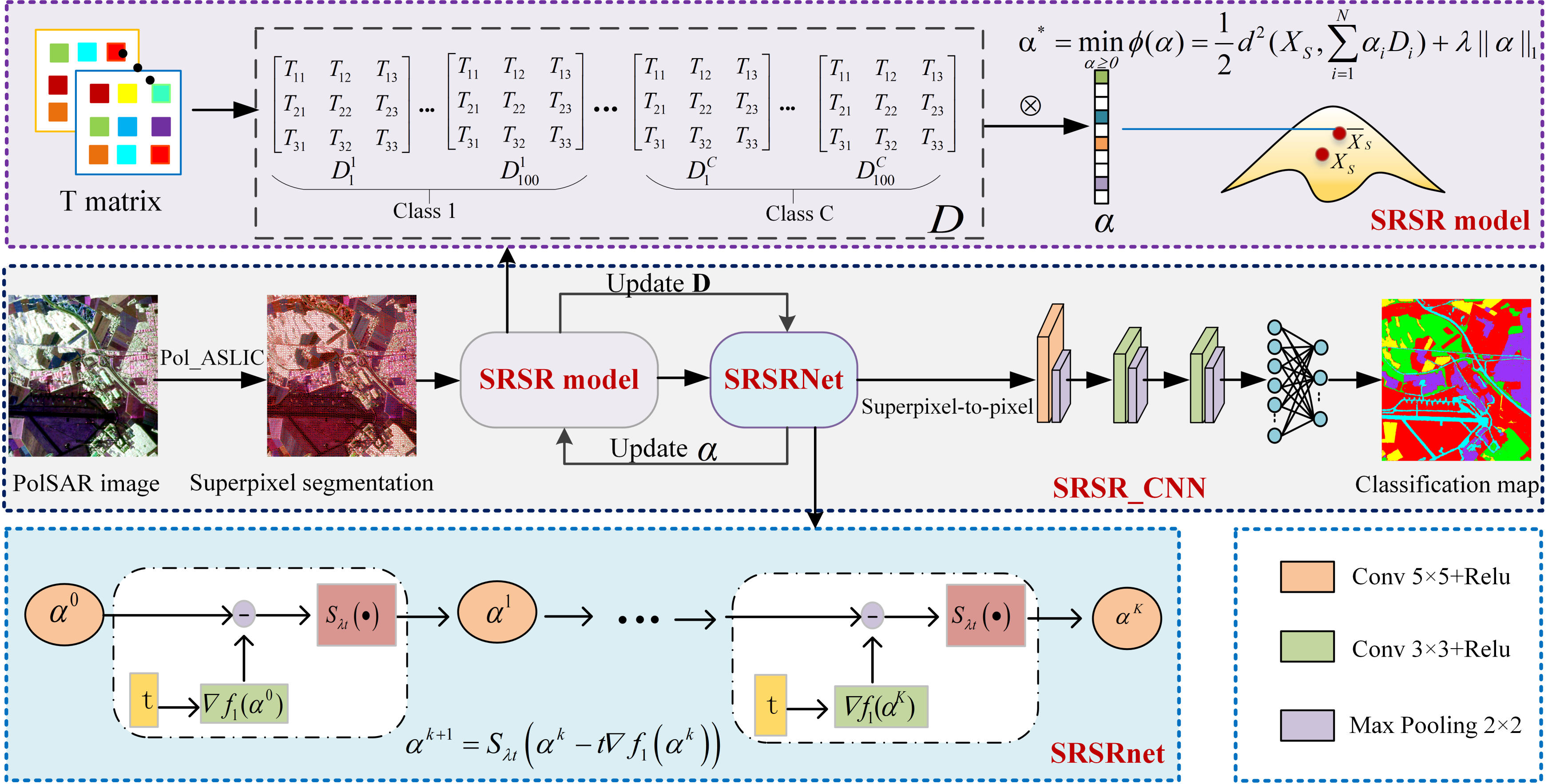}}
		\caption{The framework of the proposed SRSR\_CNN method.}
		\label{fig1}
	\end{center}
	\vskip -0.2in
\end{figure*}

In this paper, we propose a novel Riemannian sparse representation learning network (SRSR\_CNN), the framework of which is illustrated in Fig.\ref{fig1}. It primarily comprises three components: the Superpixel-based Riemannian Sparse Representation module (SRSR), SRSRNet, and CNN enhanced module. Firstly, to reduce computation complexity, a Pol\_ASLIC-based method\cite{7858788} is utilized to produce superpixels. Subsequently, a SRSR model is defined, in which a set of HPD matrix dictionary is constructed and a Riemannian metric is utilized to express the SRSR model. Secondly, an ISTA-inspired SRSRNet module is designed for learning the SRSR model, converting the SRSR optimal learning process into a network backbone. The multi-layer SRSRNet can learn automatically the dictionary and sparse coefficients, thus a sparse interpreted deep learning model is developed. Finally, a CNN-enhanced model is connected to SRSRNet to learn contextual high-level features for improving classification performance of heterogenous objects.
\vspace{-10pt}

\subsection{Superpixel-based Riemannian sparse representation model (SRSR)}

To better model an HPD matrix and mitigate the computational burden, we propose a novel superpixel-based Riemannian sparse representation (SRSR) model for PolSAR images. In the SRSR model, each superpixel is treated as a computing unit, significantly reducing the computation cost compared to pixel-wise model.
The Pol\_ASLIC method\cite{7858788} is employed to extract superpixels, and each superpixel can be represented by the average covariance matrix, defined by:
\begin{equation}\label{8}
	\footnotesize
	{X_{S}} = \frac{{\sum\limits_{i = 1}^m {{X_i}} }}{m}
\end{equation}
where $m$ is the number of pixels in superpixel $S$. ${X_i}$ is the HPD matrix of the $i$-th pixel in superpixel $S$, defined as:
\begin{equation}\label{e2}
	{X_i} = \left[ {\begin{array}{*{20}{c}}
			{{x_{11}}}&{{x_{12}}}&{{x_{13}}}\\
			{{x_{21}}}&{{x_{22}}}&{{x_{23}}}\\
			{{x_{31}}}&{{x_{32}}}&{{x_{33}}}
	\end{array}} \right]
\end{equation}
where the diagonal elements in $X_i$ are real numbers and non-diagonal elements are complex data.

Based on the generation of superpixels, we define the SRSR model as follows. Assuming $D = \left\{ {{D_1},{D_2}, \cdot  \cdot  \cdot ,{D_N}} \right\}$ is the dictionary set, in which each atom is an HPD matrix. Here, $N$ represents the number of dictionary atoms. Assuming each class has an equal number of atoms denoted as  $M$, then $N = M \times C$, where $C$ is the class number. Following the principle of sparse representation, for the average HPD matrix ${X_{S }}$ associated with each superpixel $S$, we can find an estimated value ${{\rm{\hat X}}_{\rm{S}}}$, which is a linear combination of all the dictionary atoms. This can be expressed as ${{\rm{\hat X}}_{\rm{S}}} = \alpha D = \sum\limits_{i = 1}^N {{\alpha _i}} {D_i}$, where $\alpha  = {\left[ {{\alpha _1},{\alpha _2}, \cdot  \cdot  \cdot ,{\alpha _N}} \right]^T}$ is a non-negative sparse vector. So, the SRSR model can be defined as:
\begin{equation}\label{9}
	\footnotesize
	{\alpha ^*} = \mathop {\min }\limits_{\alpha  \geqslant 0} \phi (\alpha ) = \frac{1}{2}{d^2}({X_{S }},\sum\limits_{i = 1}^N {{\alpha _i}{D_i}} ) + \lambda ||\alpha |{|_1}
\end{equation}
where the first term is the residual term, which minimizes the distance between the estimated and real values. The second term is the sparse term, which ensures that $\alpha $ is sparse.

Then, we need to choose a Riemannian metric to calculate the similarity between the two HPD matrices. Common Riemannian metrics encompass the affine invariant Riemannian metric (AIRM)\cite{RN141}, log-Euclidean Riemannian metric (LERM)
\cite{Arsigny2006LogEuclideanMF}, and Bartlett distance\cite{6378374}. Among them, the AIRM dynamically adjusts its metric based on the local geometry of the data manifold, ensuring robustness to variations and noise inherent in HPD matrices. Therefore, the AIRM metric is utilized to calculate the geometric distance in a curved manifold for PolSAR data, defined as:
\begin{equation}\label{7}
	\footnotesize
	{d_A}(X,Y) = {\left\| {\log ({X^{ - \tfrac{1}{2}}}Y{X^{ - \tfrac{1}{2}}})} \right\|_F}
\end{equation}
where $X$ and $Y$ are two covariance matrices, and ${\left\| . \right\|_F}$ is the  Frobenius norm (F-norm).
With the AIRM metric, the proposed SRSR model can be expressed as:
\begin{equation}\label{10}
	\footnotesize
	{\alpha ^*} = \mathop {\min }\limits_{\alpha  \geqslant 0} \phi (\alpha ) = \frac{1}{2}\left\| {\log (\sum\limits_{i = 1}^N {{\alpha _i}} {X_{ S }}^{ - \tfrac{1}{2}}{D_i}{X_{ S }}^{ - \tfrac{1}{2}})} \right\|_F^2 + \lambda ||\alpha |{|_1}
\end{equation}
where ${\left\| . \right\|_F}$ is the F-norm. $\lambda {\left\| . \right\|_1}$ denotes the ${l_1}$ norm-based sparse term, and $\lambda  \geqslant 0$ is the regularization parameter.

\subsection{SRSR model optimization}

The proposed SRSR model in Equ.(\ref{10}) is a non-convex optimization problem, lacking an analytical solution. To optimize the SRSR model, the iterative shrinkage-threshold algorithm (ISTA)\cite{RN140} is used, which shows high flexibility and suitability for large-scale problems. Considering Riemannian metric in the SRSR model, we re-infer the first derivative and update sparse coefficients and dictionary for Equ.(\ref{10}).

1) \textbf{Inferring first derivative}: We divide the Equ.(\ref{10}) into two terms: the first term represents the residual term, while the second term denotes the sparse term, defined as:
\begin{equation}\label{11}
	\footnotesize
	\begin{aligned}
		\min f\left( \alpha  \right) &= \frac{1}{2}\left\| {\log (\sum\limits_{i = 1}^N {{\alpha _i}} {X_{ S }}^{ - \tfrac{1}{2}}{D_i}{X_{ S }}^{ - \tfrac{1}{2}})} \right\|_F^2 + \lambda ||\alpha |{|_1} \\
		&= {f_1}\left( \alpha  \right) + {f_2}\left( \alpha  \right)
	\end{aligned}
\end{equation}

Assuming $M\left( \alpha  \right) = \sum\limits_{j = 1}^n {{\alpha _j}} {D_j}$ and $H = {X_{ S }}^{ - \frac{1}{2}}$. According to the definition of the F-norm and the chain principle of calculus, the first-order derivative of the first term is inferred as:
\begin{equation}\label{12}
	\footnotesize
	f_1^\prime \left( \alpha  \right) = Tr\left( {\log \left( {HM\left( \alpha  \right)H} \right){{\left( {HM\left( \alpha  \right)H} \right)}^{ - 1}}H{M^\prime }\left( \alpha  \right)H} \right)
\end{equation}

Then, letting $M\left( {{\alpha _p}} \right) = {\alpha _p}{D_p} + \sum\limits_{i \ne p} {{\alpha _i}{D_i}} $, $p \in \left\{ {1,2, \cdot  \cdot  \cdot ,N} \right\}$ is the $p$-th element of coefficient $\alpha $, then
\begin{equation}\label{13}
	\footnotesize
	\frac{{\partial {f_1}\left( \alpha  \right)}}{{\partial {\alpha _p}}} = Tr\left( {\log \left( {HM\left( {{\alpha _p}} \right)H} \right){{\left( {HM\left( {{\alpha _p}} \right)H} \right)}^{ - 1}}H{D_p}H} \right)
\end{equation}

Besides, the first-order derivative of the second term is simple to infer by:
\begin{equation}\label{14}
	\footnotesize
	{f_2}^\prime \left( \alpha  \right) = \lambda \operatorname{sgn} \left( \alpha  \right)
\end{equation}

Finally, we can obtain the first derivative of the objective function in Equ.(\ref{11}), which can be represented as:
\begin{equation}\label{15}
	\scriptsize
	\begin{aligned}
		\frac{{\partial f\left( \alpha  \right)}}{{\partial {\alpha _p}}} &= Tr\left( {\log \left( {HM\left( {{\alpha _p}} \right)H} \right){{\left( {HM\left( {{\alpha _p}} \right)H} \right)}^{ - 1}}H{D_p}H} \right) + \lambda \operatorname{sgn} \left( \alpha  \right)
	\end{aligned}
\end{equation}

2) \textbf{Iteration update sparse coefficients}: After obtaining the first derivative of the objective function, based on the ISTA algorithm\cite{RN140}, the sparse coefficients $\alpha$ can be iterated using the following formulation:
\begin{equation}\label{18}
	\footnotesize
	{\alpha ^{k + 1}} = {S_{\lambda t}}\left( {{\alpha ^k} - t\nabla {f_1}\left( {{\alpha ^k}} \right)} \right)
\end{equation}
where ${S_{\lambda t}}\left(  \cdot  \right)$ is the soft threshold operation. ${\alpha ^k}$ and ${\alpha ^k+1}$ are the sparse coefficients in the $k$-th and $k+1$-th iterations. $t$ is the step factor. Replacing $\nabla {f_1}\left( {{\alpha ^k}} \right)$ by Equ.(\ref{13}), Equ.(\ref{18}) can be written as:
\begin{equation}\label{19}
	\footnotesize
	{\alpha ^{k + 1}} = {S_{\lambda t}}({\alpha ^k} - tTr(\log (HM({\alpha _p})H){(HM({\alpha _p})H)^{ - 1}}H{D_p}H))
\end{equation}

3) \textbf{Iteration update dictionary atoms}: To make the SRSR model more flexible, we fix $\alpha$ and update the dictionary atoms in each iteration. Since the dictionary atoms are situated in Riemannian space instead of Euclidean space, a Riemannian dictionary learning method\cite{7565529} is utilized to perform dictionary learning. The derivation process is provided in Appendix A. Given ${\alpha ^k}$ and ${D^k}$, The updated dictionary is defined as:
\begin{equation}\label{20}
	\scriptsize
	{D^*} = \min \theta (D) = \frac{1}{2}\left\| {\log (\sum\limits_{i = 1}^N {{\alpha _i}} {X_{ S }}^{ - \tfrac{1}{2}}{D_i}{X_{S }}^{ - \tfrac{1}{2}})} \right\|_F^2 + {\lambda_B}\sum\limits_{i = 1}^N {Tr\left( {{D_i}} \right)}
\end{equation}
\vspace{-10pt}
\subsection{SRSRNet design}

The iterative solution of sparse coefficients, as described in Equ.(\ref{19}), involves computationally expensive matrix operations. Fortunately, the ISTA algorithm iteration process bears resemblance to the convolution operation in a network\cite{RN144,10231373}. Inspired by the ISTAnet, we develop a Superpixel-based Riemannian Sparse Representation Network (SRSRNet), illustrated in Fig.\ref{fig1}. The SRSRNet is designed by unfolding the iterative procedure in Equ.(\ref{18}) as a network backbone, and multiple layers network can solve the sparse coefficients. The SRSRnet can not only learn features in the network module, but also make the network interpretable with mathematical model-based design. The proposed SRSRNet iteration procedure is outlined as follows.

\textbf{Network Initialization:} Firstly, we initialize the dictionary atom ${D^0}$ and sparse coefficient ${\alpha ^0}$. In this paper, the initial dictionary atoms are randomly chosen from the labeled samples for each class, and then ${D^0} = \left\{ {{D_1},{D_2}, \cdot  \cdot  \cdot ,{D_N}} \right\},N = M \times C$. To obtain reasonable ${\alpha ^0}$, the initial coefficient ${\alpha ^0}$ can be derived from the initial dictionary ${D^0}$ using the spectral projected gradient (SPG)\cite{RN149} algorithm. Then, the initial first derivative $\nabla f\left( {{\alpha ^0}} \right)$ is calculated by Equ.(\ref{15}).

\textbf{SRSRNet Iteration:}

\begin{figure}[ht]
	\begin{center}
		\centerline{\includegraphics[height=0.08\textheight]{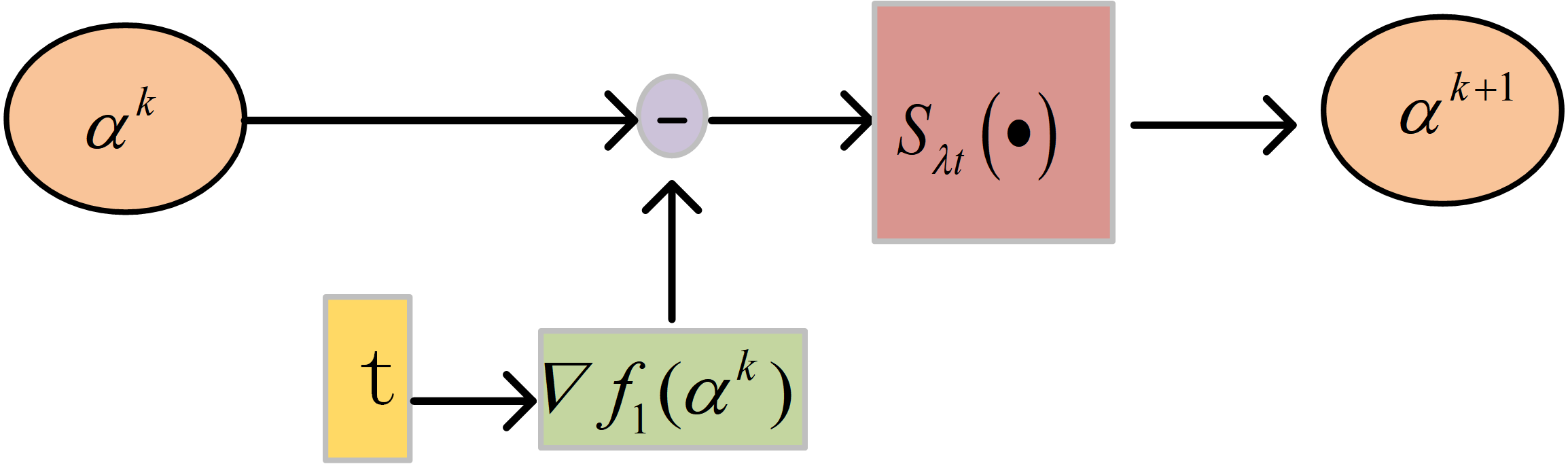}}
		\caption{A single layer obtained by unfolding Eq.(\ref{18}).}
		\label{fig2}
	\end{center}
\end{figure}

(1) \emph{Update sparse coefficient:} The iterative equation in (\ref{18}) can be unfolded into a SRSRNet backbone shown in Fig.\ref{fig2}. The process of constructing the network backbone in Fig.\ref{fig2} involves three steps. First, we compute ${\nabla {f_1}\left( {{\alpha ^k}} \right)}$, which represents the first derivative in Equ.(\ref{13}). ${\alpha ^k}$ denotes the output of the $k$-th layer. Second, we multiply ${\nabla {f_1}\left( {{\alpha ^k}} \right)}$ by the step size $t$ and subtract the result from ${\alpha ^k}$. Finally, we apply the soft-thresholding operator ${S_{\lambda t}}\left( \cdot \right)$, functioning as a non-linear ReLU layer, to obtain the output of the ($k+1$)-th layer ${\alpha ^{k+1}}$.
\vspace{-10pt}

(2) \emph{Update dictionary atom:}
After fixing ${\alpha ^{k+1}}$, we update the dictionary atoms by Equ.(\ref{20}) using the Riemannian Conjugate Gradient (CG) method\cite{7565529}. Following the first iteration, the coefficients and dictionary are updated, equivalent to the first layer of the SRSRNet backbone.

Through multiple layers of the SRSRNet, the coefficients and dictionary can converge towards the estimated optimal solution. The proposed SRSRNet structure is illustrated in the bottom of Fig.\ref{fig1}.

\subsection{CNN-enhanced module}

The SRSRNet excels at learning geometric features of covariance matrices in Riemannian space with the guidance of SRSR model, while it lacks higher-level contextual features. To learn high-level semantic features, we augment the SRSRNet with a multiple-layer CNN model. Firstly, the superpixel-wise features $F$ should be projected into pixels' features $F'$. Then, the high-level features $H$ is obtained by multi-layer CNN. The feature learning procedure can be expressed as:
\begin{equation}\label{23}
	\footnotesize
	H = CNN\left( {\textbf{P}\left( {SRSRNet\left( \textbf{C} \right)} \right)} \right)
\end{equation}
where $\textbf{C}$ is the HPD matrix, and $\textbf{P}$ is the feature projection operation. In this paper, a three-layer CNN is applied to extract contextual features, followed by a softmax layer for classification. The whole algorithm procedure of the proposed SRSR\_CNN is given in \textbf{Algorithm 1}.

\begin{algorithm*}[tb]
\small
	\caption{Algorithm procedure of the proposed SRSR\_CNN method}
	\label{Algorithm 1}
	\begin{algorithmic}
		\STATE {\bfseries Input:} PolSAR covariance matrix $\textbf{C}$, PolSAR PauliRGB image $\textbf{I}$, and label map $L$. Superpixels scale parameter factor $\delta $, step size factor $t$ and the regularization parameter $\lambda $.
		\STATE {\bfseries Output:} Classification map $\textbf{Y}$.
		
		\emph{Step 1:} Using the Pol\_ASLIC method, obtain the superpixels $S$ from PolSAR image \textbf{I}.\\
		\emph{Step 2:} Compute the average covariance matrix ${X_{ s }}$ for each superpixel $s$ from PolSAR covariance matrix $\textbf{C}$.\\
		\emph{Step 3:} Initialize the dictionary atom ${D^0}$ by randomly selecting 100 pixels from each class of labeled pixels and initialize the sparse coefficient ${\alpha ^0}$ by SPG algorithm.\\
		\emph{Step 4:} Update sparse coefficient and dictionary atoms by equations(\ref{18}) and (\ref{20}).\\
		\emph{Step 5:} Obtain final superpixel sparse representation features $F$ by the SRSRNet.\\
		\emph{Step 6:} Project superpixels' features $F$ to pixels' features $F'$.\\
		\emph{Step 7:} Learn high-level features $H$ by CNN model with Equ.(\ref{23}).\\
		\emph{Step 8:} Obtain the final classification map $\textbf{Y}$ by the softmax classifier.
	\end{algorithmic}
\end{algorithm*}

\section{Experiments}

\subsection{Experimental data and settings}

To evaluate the effectiveness of the proposed SRSR\_CNN method, we select three real PolSAR data sets from different bands and sensors, which are described in detail as follows.

\begin{figure*}
	\centering
	\setlength{\fboxrule}{0.2pt}
	\setlength{\fboxsep}{0.01mm}
	\includegraphics[height=0.23\textheight]{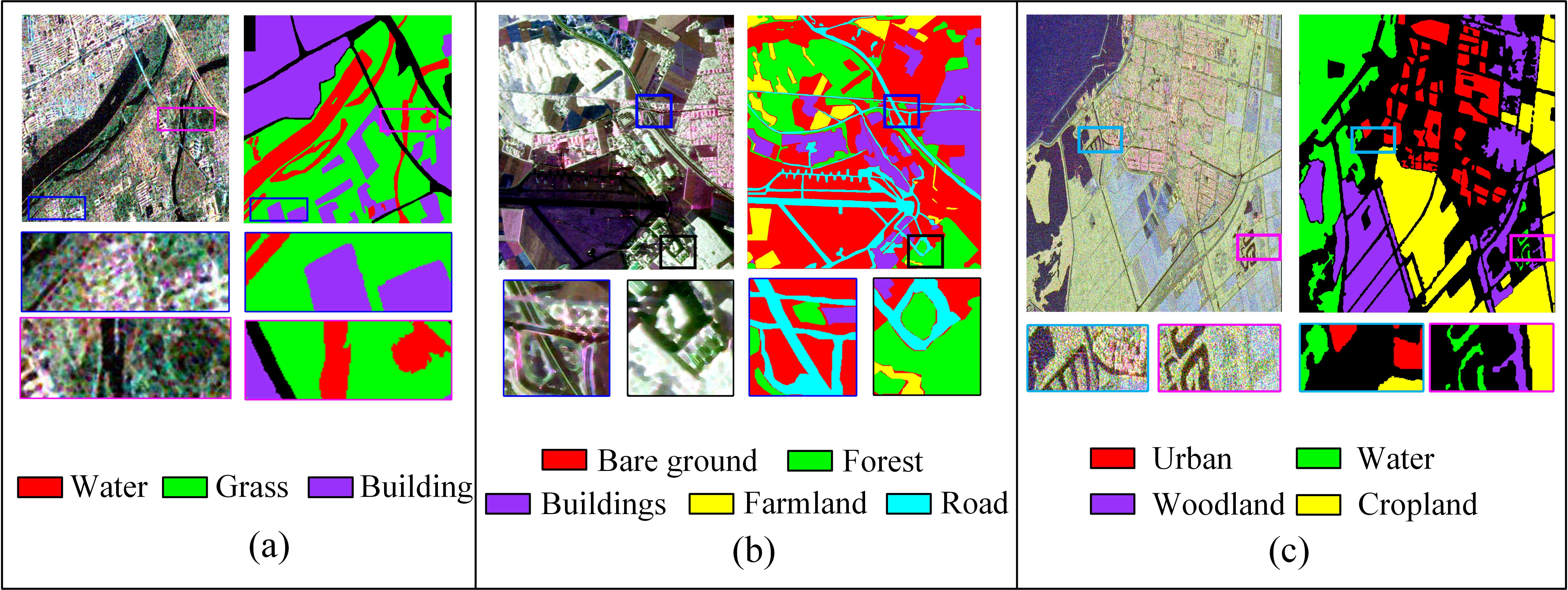}
	\caption{PauliRGB images and the ground truth maps on the three considered datasets, i.e., Area of Xi'an, Oberpfaffenhofen, and Flevoland.}
	\label{fig4}
\end{figure*}

A)\emph{\textbf{Xi'an data set}}: This is a C band full PolSAR image from Xi'an area obtained by RADARSAT-2 system. The spatial resolution is $8\times8$m, with an image size of $512\times512$ pixels. There are mainly three different types of land covers, including \emph{water}, \emph{grass} and \emph{building}. The PauliRGB image and its corresponding label map are shown in Fig.\ref{fig4} (a). The black area in the label map is the unlabeled pixels, which are not involved during the accuracy computation. It's the same definition for the following PolSAR data sets.

B)\emph{\textbf{Oberpfaffenhofen data set}}: This is an E-SAR L band full PolSAR data set covering the Oberpfaffenhofen area. It is caught from German Aerospace Center with the resolution of $3\times2.2$m. Its size is $1300\times1200$ pixels and it primarily consists of five classes including \emph{bare ground}, \emph{forest}, \emph{buildings}, \emph{farmland} and \emph{road}. Figure \ref{fig4} (b) exhibits the PauliRGB image and ground truth map.

C)\emph{\textbf{Flevoland data set}}: The last image depicts the Flevoland area in the C band from RADARSAT-2 sensor. The pixel size is $1400\times1200$ pixel and its resolution is $12\times8$m. There are \emph{urban}, \emph{water}, \emph{woodland} and \emph{cropland} in this scene. Figure \ref{fig4}(c) illustrates the PauliRGB image and the ground truth map.

In the experiments, we randomly select 100 pixels for each category from the covariance matrices to serve as dictionary atoms. The SRSRNet is unfolded with four layers, and the regularization parameter $\lambda$ and gradient descent step size $t$ are set to 0.5 and 0.0001, respectively. The CNN module is configured with three layers. Other parameter settings include a learning rate of 0.001, a patch size of $9\times9$, and a batch size of 128 with 50 epochs. The training sample ratio is set to 10\%, with the remaining 90\% for testing. The experimental setup includes a Windows 10 operating system and an Intel(R) Core(TM) i7-10700 CPU. Additionally, it features an NVIDIA GeForce RTX 3060 GPU and 64GB of RAM. The programming environment is Python 3.7 with PyTorch GPU 1.12.1.

To fairly assess classification performance of the proposed method, we choose six classification algorithms for comparison, including a representation learning based method (KNNRS\cite{RN146}), CNN based variants (CVCNN\cite{RN86}, 3DCNN\cite{RN135}, and PolMPCNN\cite{9424197}), and GCN based variants (DFGCN\cite{RN147} and CEGCN\cite{10281574}). For optimal performance, all compared methods use the same parameters as their original papers. To assess and compare the classification performance of different methods and modules, we employ six commonly evaluation indicators, including: user's accuracy (UA), overall accuracy (OA), average accuracy (AA), Kappa coefficient, F1\_Score, and MIoU.

\subsection{Experimental results on Xi'an data set}
\begin{figure*}
	\centering
	\setlength{\fboxrule}{0.2pt}
	\setlength{\fboxsep}{0.01mm}
	\subfloat[]{\includegraphics[height=0.17\textheight]{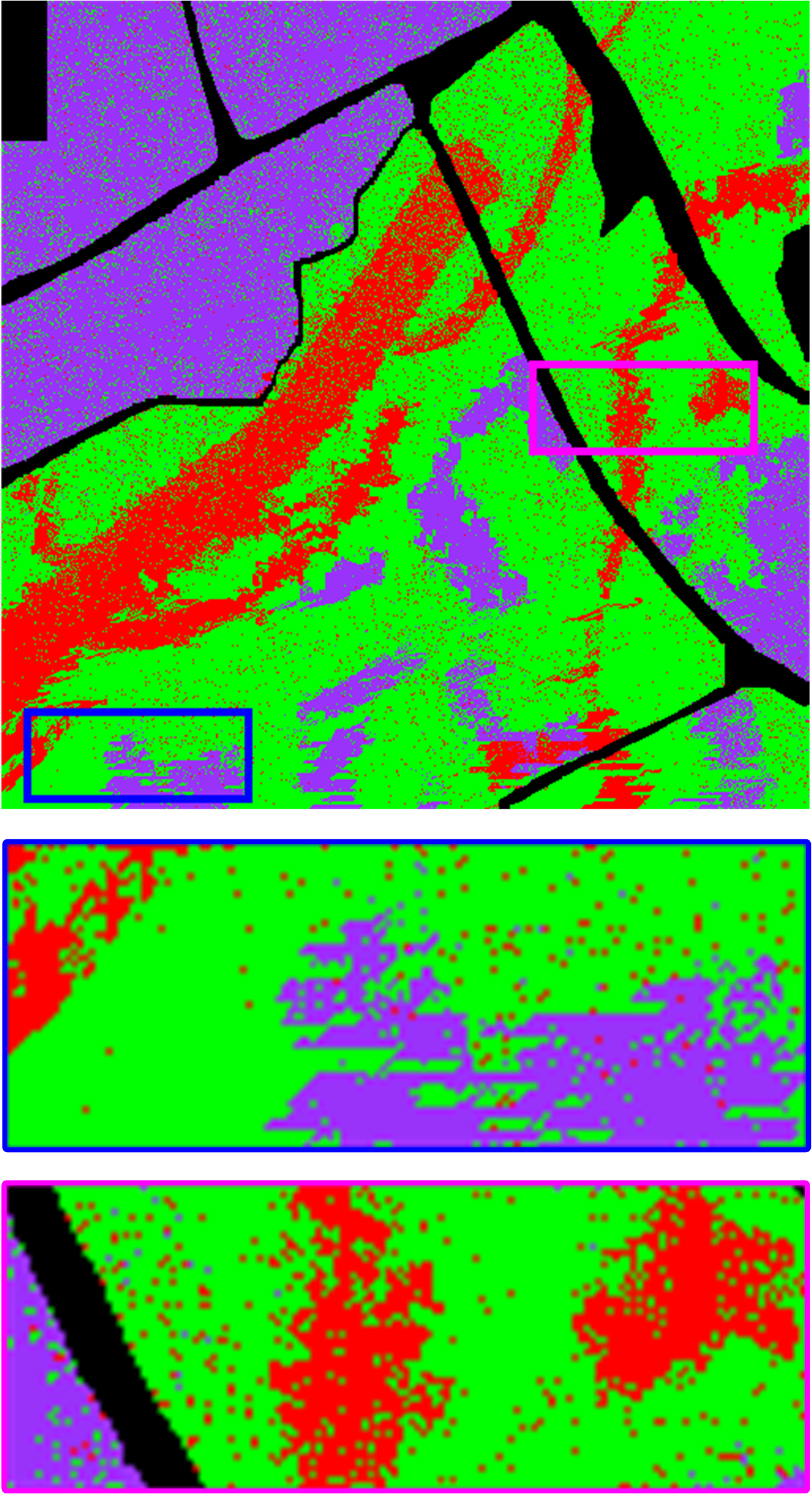}}
	\hspace{0.2mm}
	\subfloat[]{\includegraphics[height=0.17\textheight]{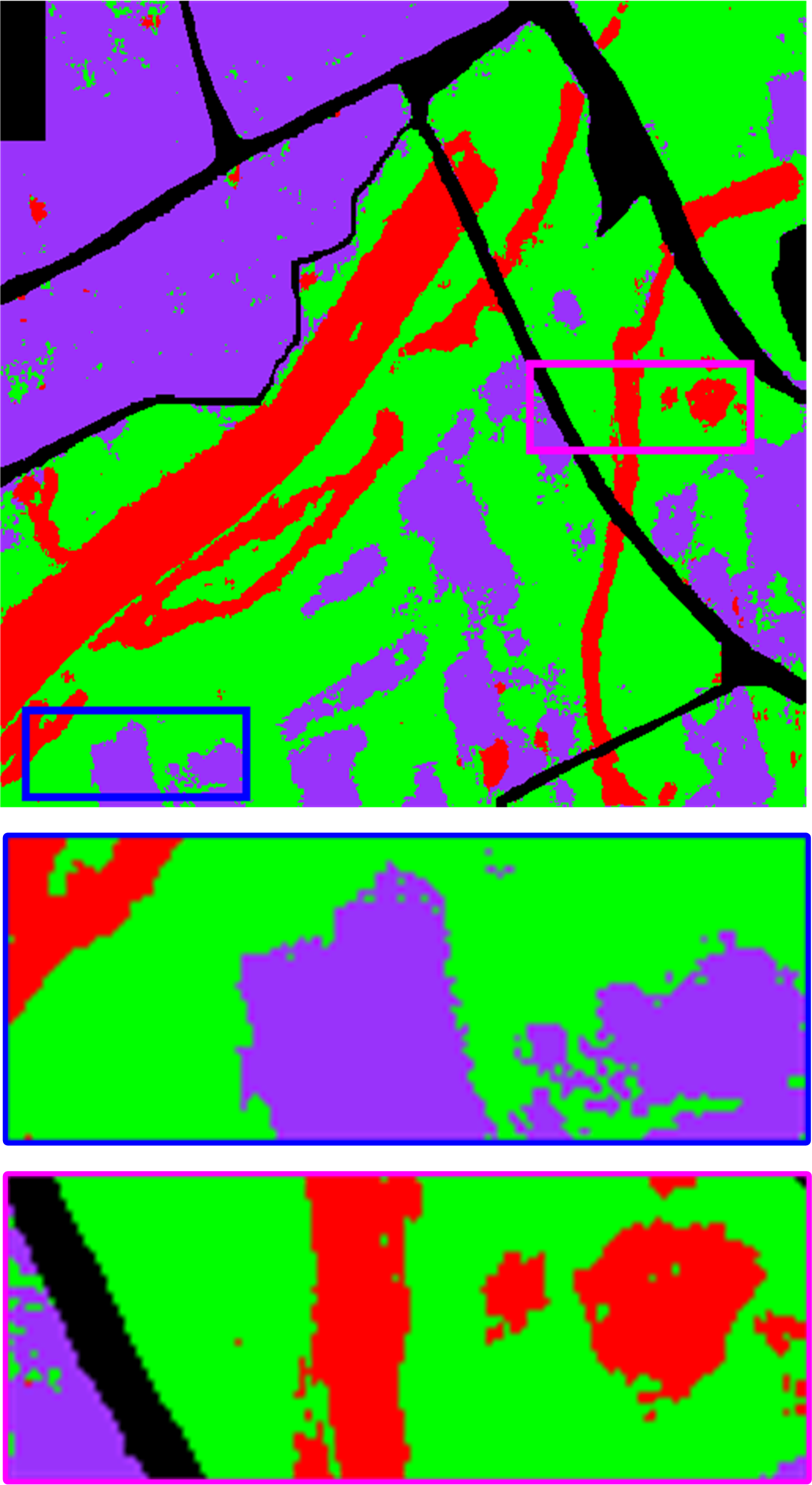}}
	\hspace{0.2mm}
	\subfloat[]{\includegraphics[height=0.17\textheight]{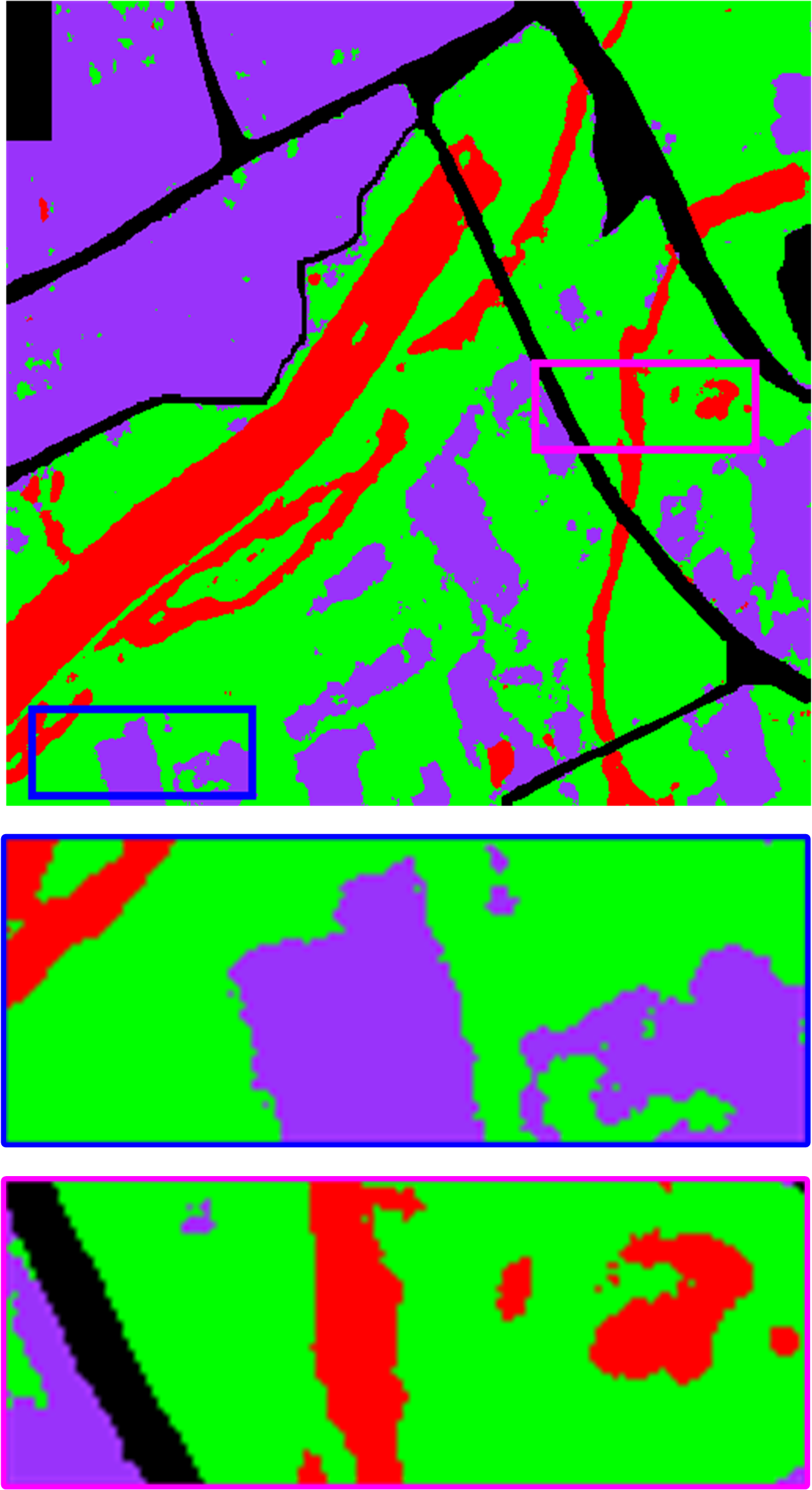}}
    \hspace{0.2mm}
	\subfloat[]{\includegraphics[height=0.17\textheight]{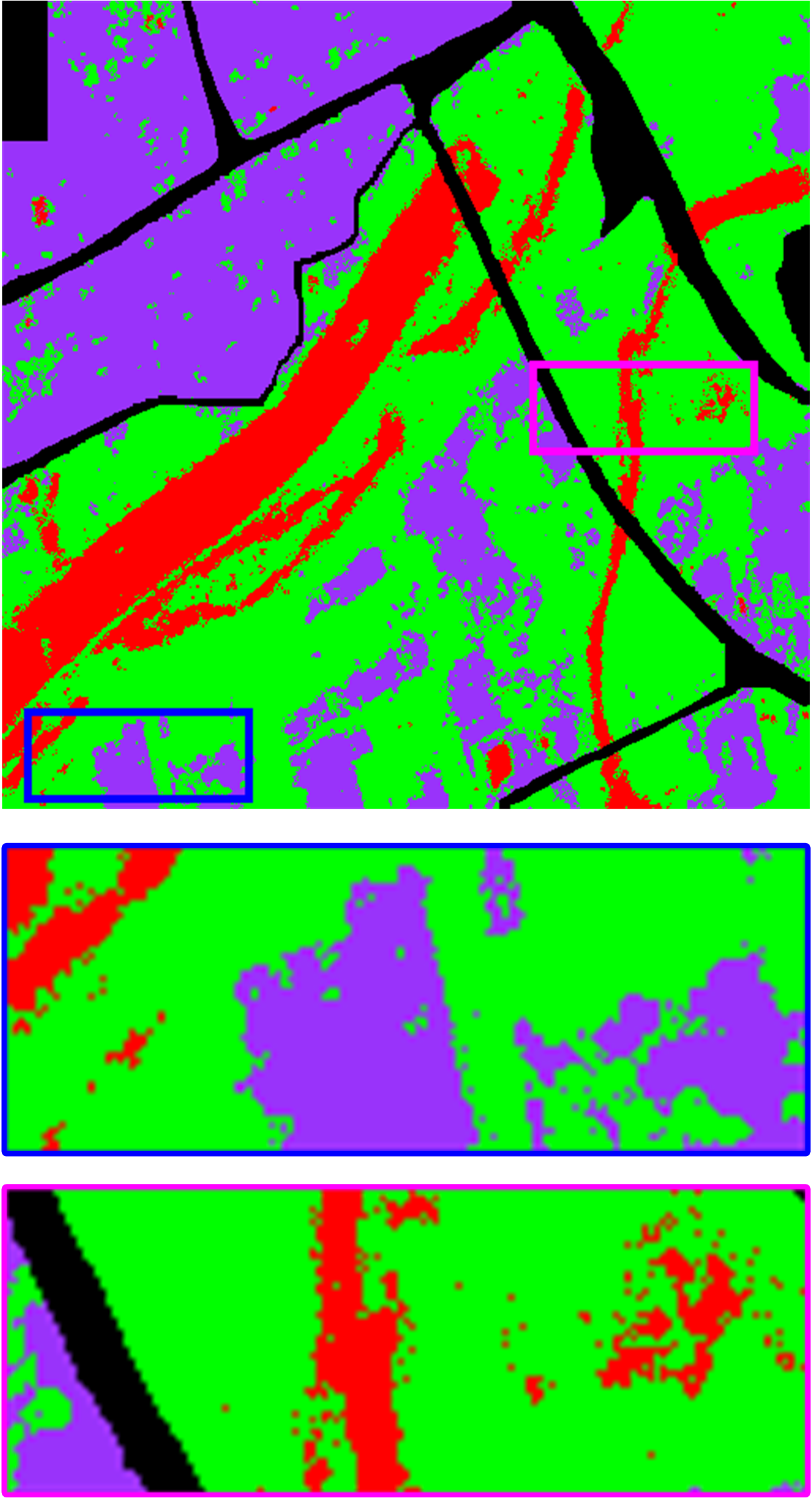}}
	\hspace{0.2mm}
	\subfloat[]{\includegraphics[height=0.17\textheight]{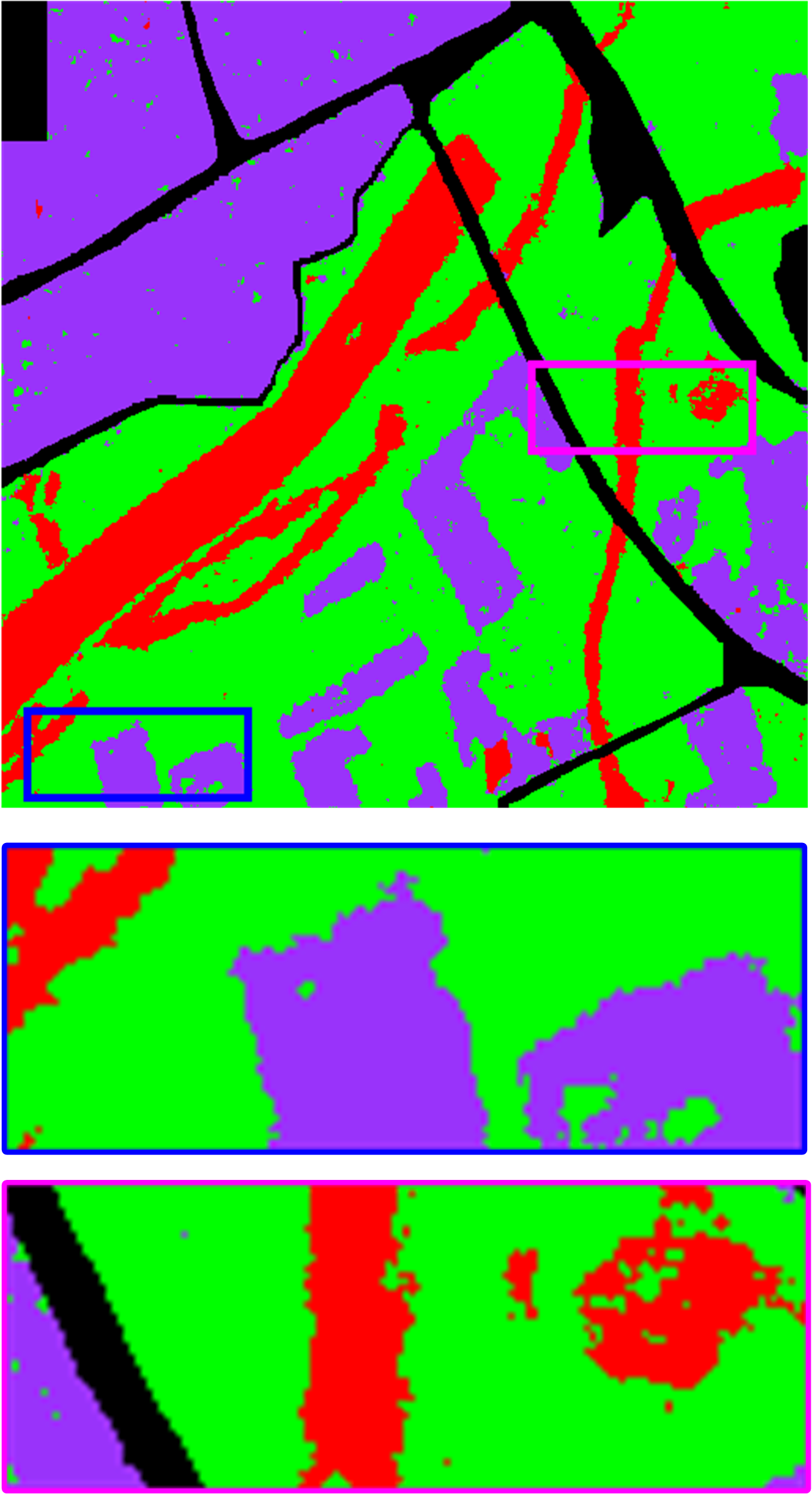}}
	\hspace{0.2mm}
	\subfloat[]{\includegraphics[height=0.17\textheight]{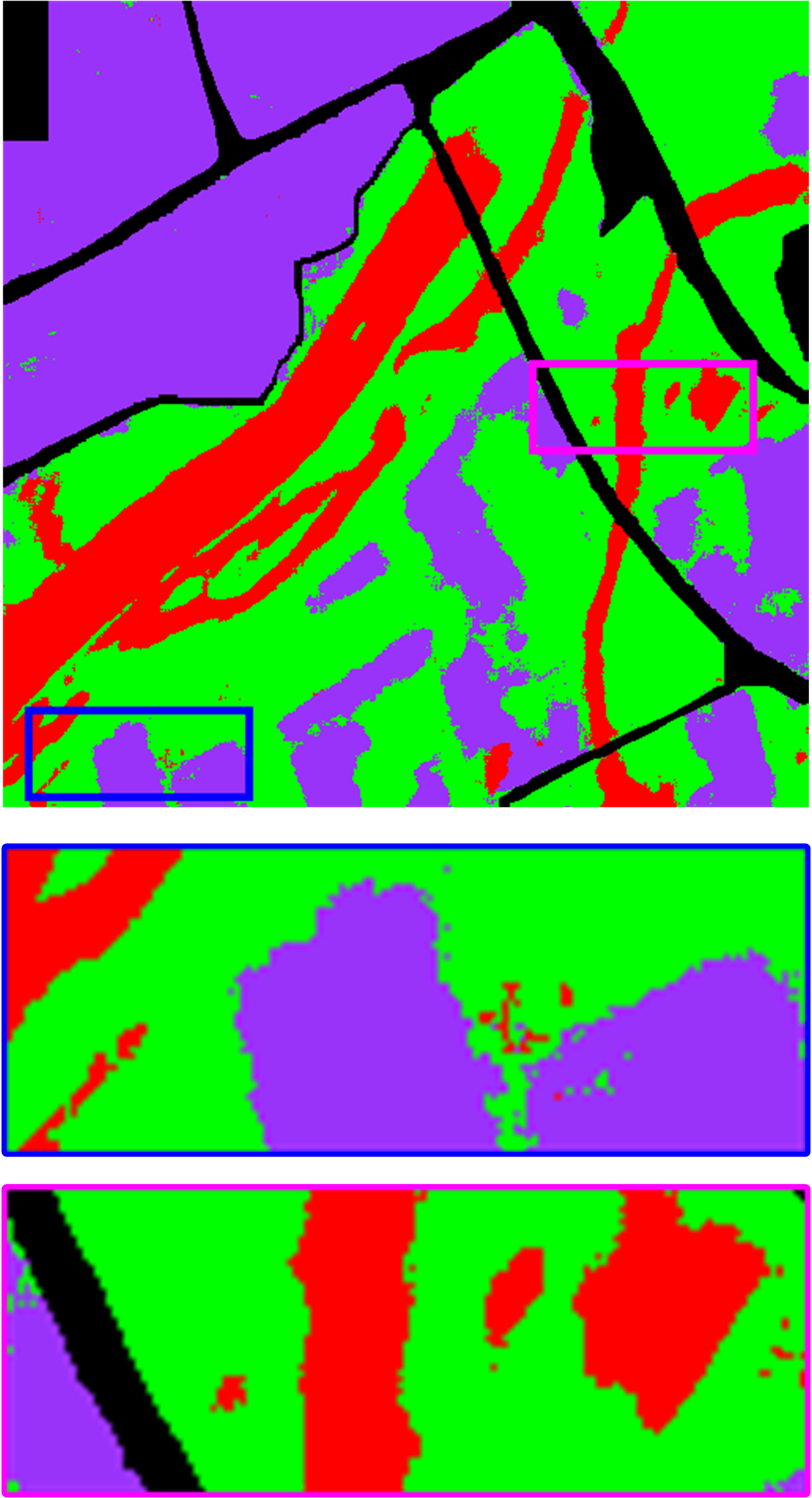}}
	\hspace{0.2mm}
	\subfloat[]{\includegraphics[height=0.17\textheight]{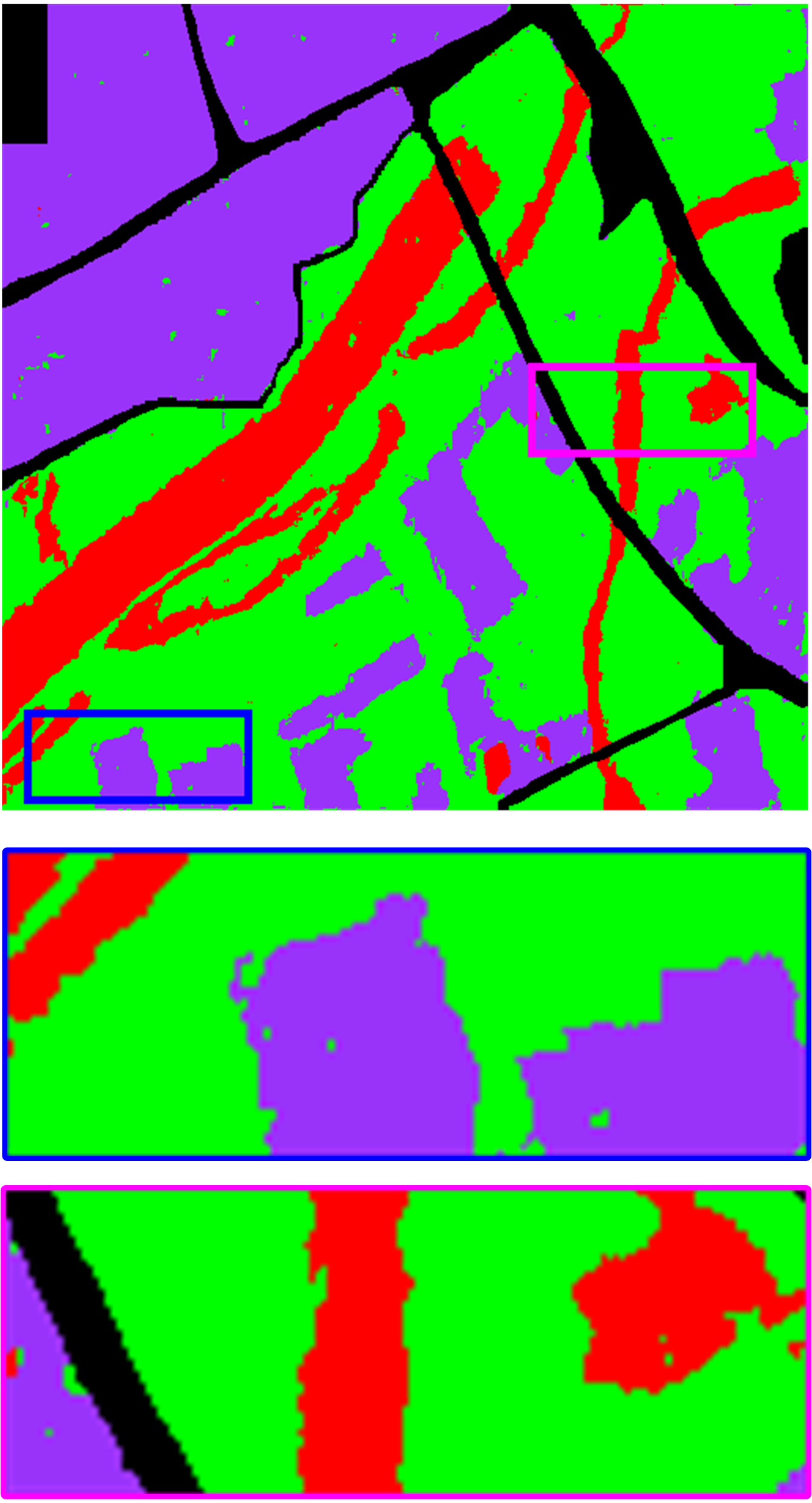}}
	
	\includegraphics[height=0.02\textheight]{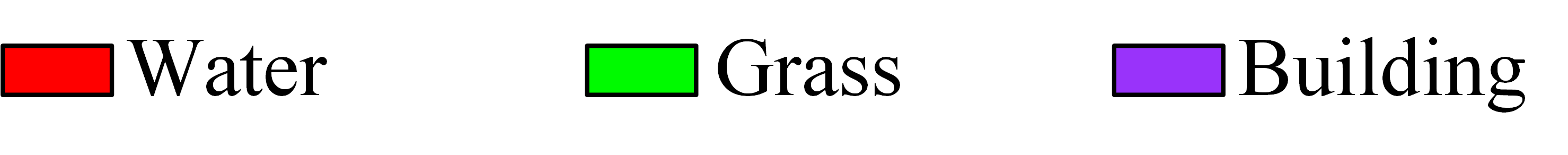}
	\caption{Classification results of the Xi'an data set. (a) KNNRS; (b) CVCNN; (c) 3DCNN; (d) DFGCN; (e) CEGCN; (f) PolMPCNN; (g) SRSR\_CNN.}
	\label{fig5}
	\end{figure*}

The classification results of the six compared methods, and the proposed SRSR\_CNN are depicted in Figs.\ref{fig5}(a)-(g), respectively. Upon analyzing the Fig.\ref{fig5}(a), it is evident that the result of KNNRS contains a significant amount of noise. The CVCNN in (b) can obviously reduce noises by utilizing the amplitude and phase information of the PolSAR images. However, some pixels in the \emph{building} class are misclassified as \emph{water} and \emph{grass}. The 3DCNN in (c) has finer boundaries in \emph{water} class, but there are some misclassified pixels in the \emph{building}. The pixel-level DFGCN limits its ability to capture global information, resulting in poor classification performance in heterogeneous regions with complex structures. Moreover, the CEGCN in (e) exhibits numerous pixels in \emph{building} class due to its exclusive reliance on global information. The PolMPCNN in (f) achieves excellent classification performance in both homogeneous and heterogeneous regions, while the boundary pixels of \emph{building} are misclassified into \emph{grass}. The proposed SRSR\_CNN involves projecting superpixel features into pixels and utilizing the CNN to extract deep features, thereby effectively enhancing the classification accuracy.

	\begin{table*}[ht]
		\footnotesize
		\begin{center}
			\caption
			{ \label{t1}
				Classification accuracy of different methods on Xi'an Data Set(\%).}
			\begin{tabular}{p{2.2cm}|p{1.3cm}p{1.3cm}p{1.3cm}p{1.3cm}p{1.3cm}p{1.3cm}p{1.3cm}}
				\hline
				class&KNNRS&CVCNN&3DCNN&DFGCN&CEGCN&PolMPCNN&proposed\\
				\hline
				water&82.70&94.55&90.27&84.64&94.47&\textbf{95.52}&94.09\\
				grass&89.86&90.68&93.60&91.79&96.50&90.95&\textbf{97.22}\\
				building&80.47&93.81&93.91&87.21&96.54&\textbf{97.68}&97.51\\
				\hline
				OA&85.47&92.37&93.21&89.10&96.21&94.01&\textbf{96.98}\\
				AA&84.34&93.01&92.60&87.88&95.83&94.71&\textbf{96.55}\\
				Kappa&75.99&87.51&88.77&81.84&93.74&90.25&\textbf{95.01}\\
				F1\_Score&83.81&92.01&92.87&88.76&95.80&93.24&\textbf{96.62}\\
				MIoU&72.46&85.22&86.70&79.80&91.97&87.45&\textbf{93.48}\\
				
				\hline
			\end{tabular}
		\end{center}
	\end{table*}

	Table \ref{t1} gives the classification accuracy of the compared and proposed methods. Compared with other methods, the proposed approach demonstrates the highest classification accuracies in all the OA, AA, Kappa coefficient, F1\_Score, and MIoU. The accuracy of KNNRS is below 90\% in three categories. What's more, it exhibits the lowest classification accuracy at only 80.47\% in \emph{building} due to its limited effectiveness in classifying heterogeneous regions. Better classification performance appears in both CVCNN and 3DCNN methods with over 90\%. However, the CVCNN tends to produce misclassifications, particularly in the \emph{grass} class. In addition, there is misclassification for 3DCNN and DFGCN in \emph{water}. The CEGCN can obtain the superior performance in \emph{grass} class, while it still produces false classes, especially in \emph{water}. In contrast, the PolMPCNN achieves the highest accuracies of 95.52\% and 97.68\% in the \emph{water} and \emph{building} classes by leveraging multi-scale convolution to effectively capture local and global information. The proposed method utilizes SRSRNet to capture covariance matrix correlations and then employs CNN to extract high-level features, achieving a substantial improvement in classification accuracy.
	
	\subsection{Experimental results on Oberpfaffenhofen data set}
\begin{figure*}
		\centering
		\setlength{\fboxrule}{0.2pt}
		\setlength{\fboxsep}{0.01mm} 
\subfloat[]{\includegraphics[height=0.16\textheight]{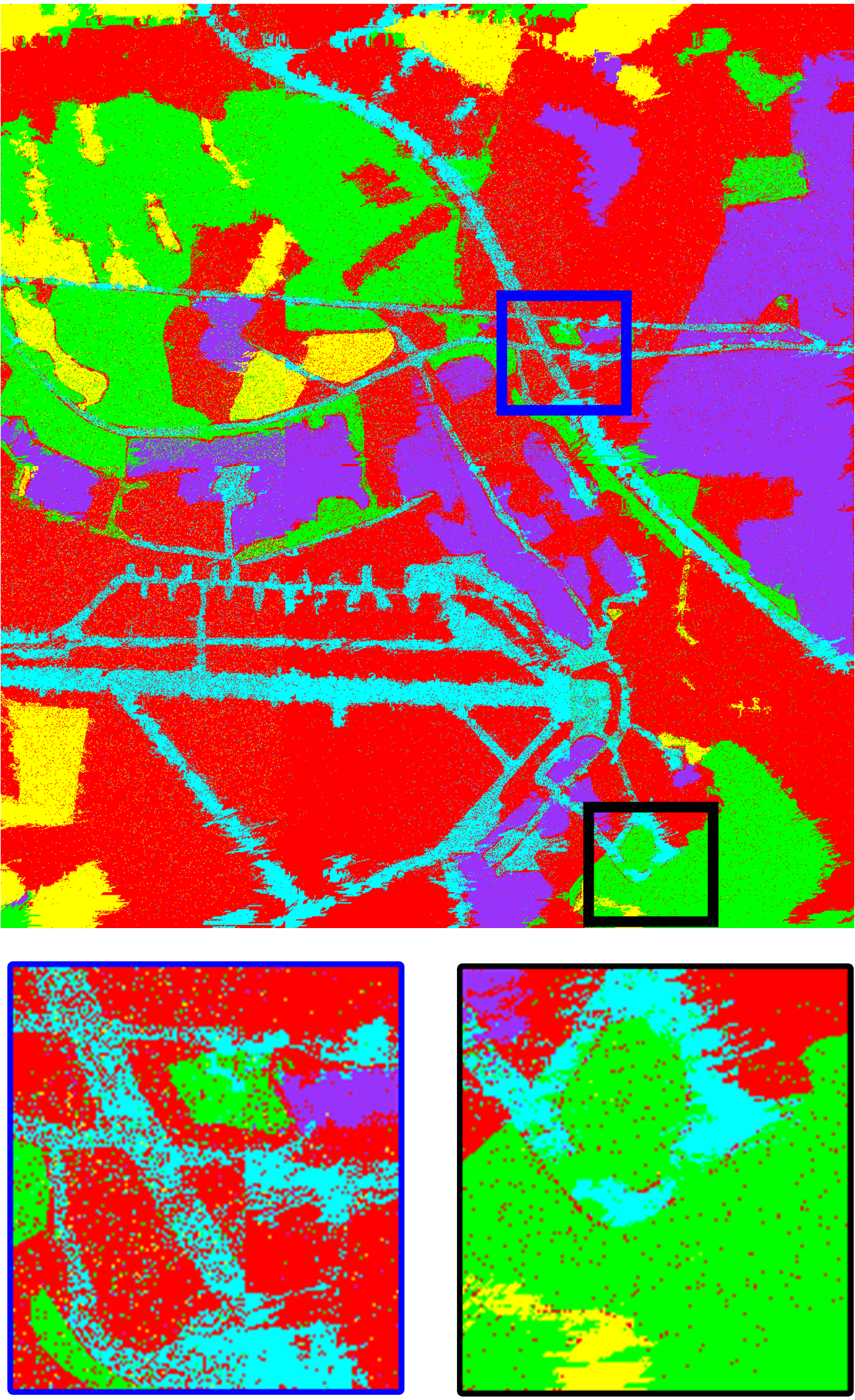}}
		\hspace{0.2mm}		\subfloat[]{\includegraphics[height=0.16\textheight]{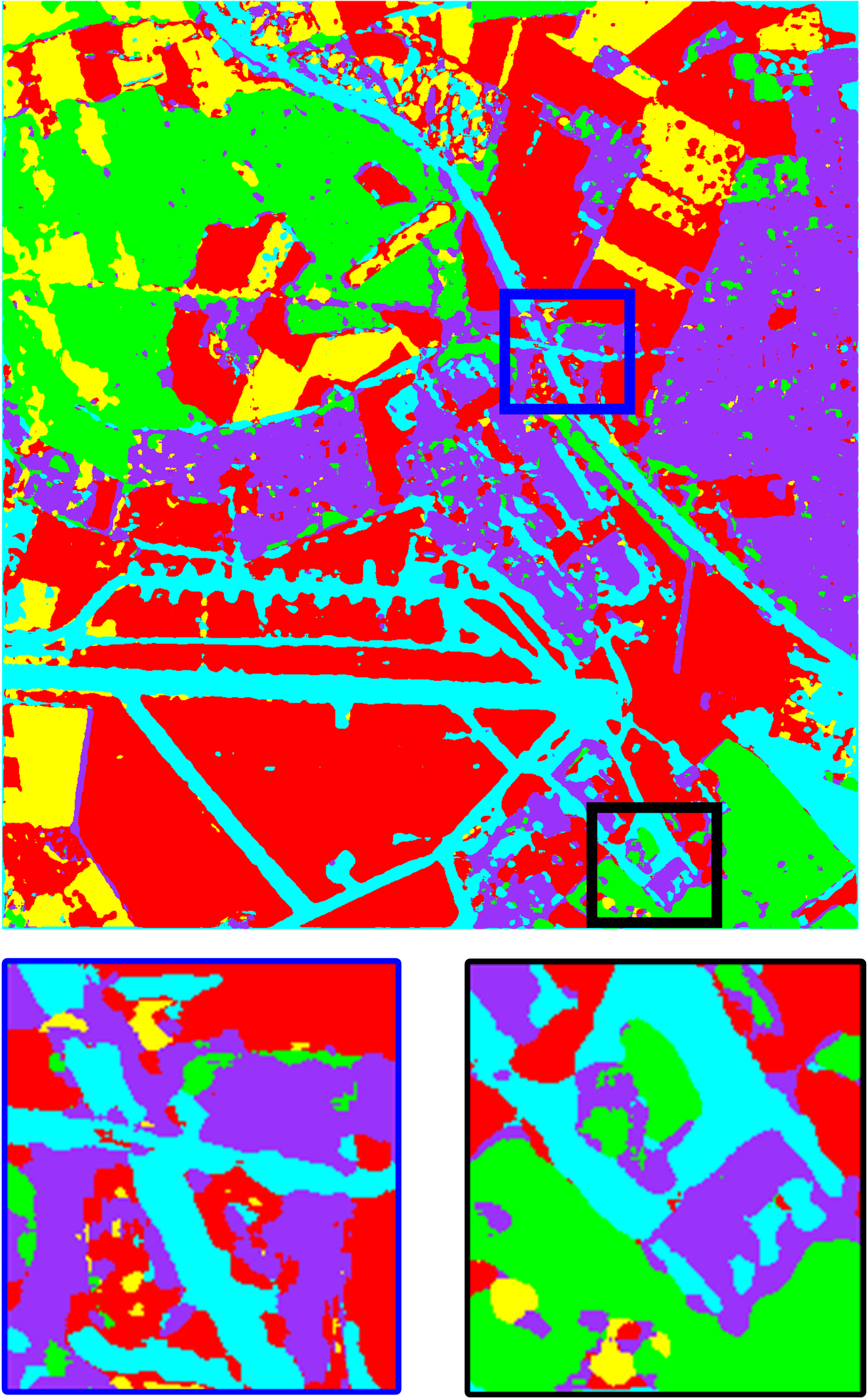}}
		\hspace{0.2mm}		\subfloat[]{\includegraphics[height=0.16\textheight]{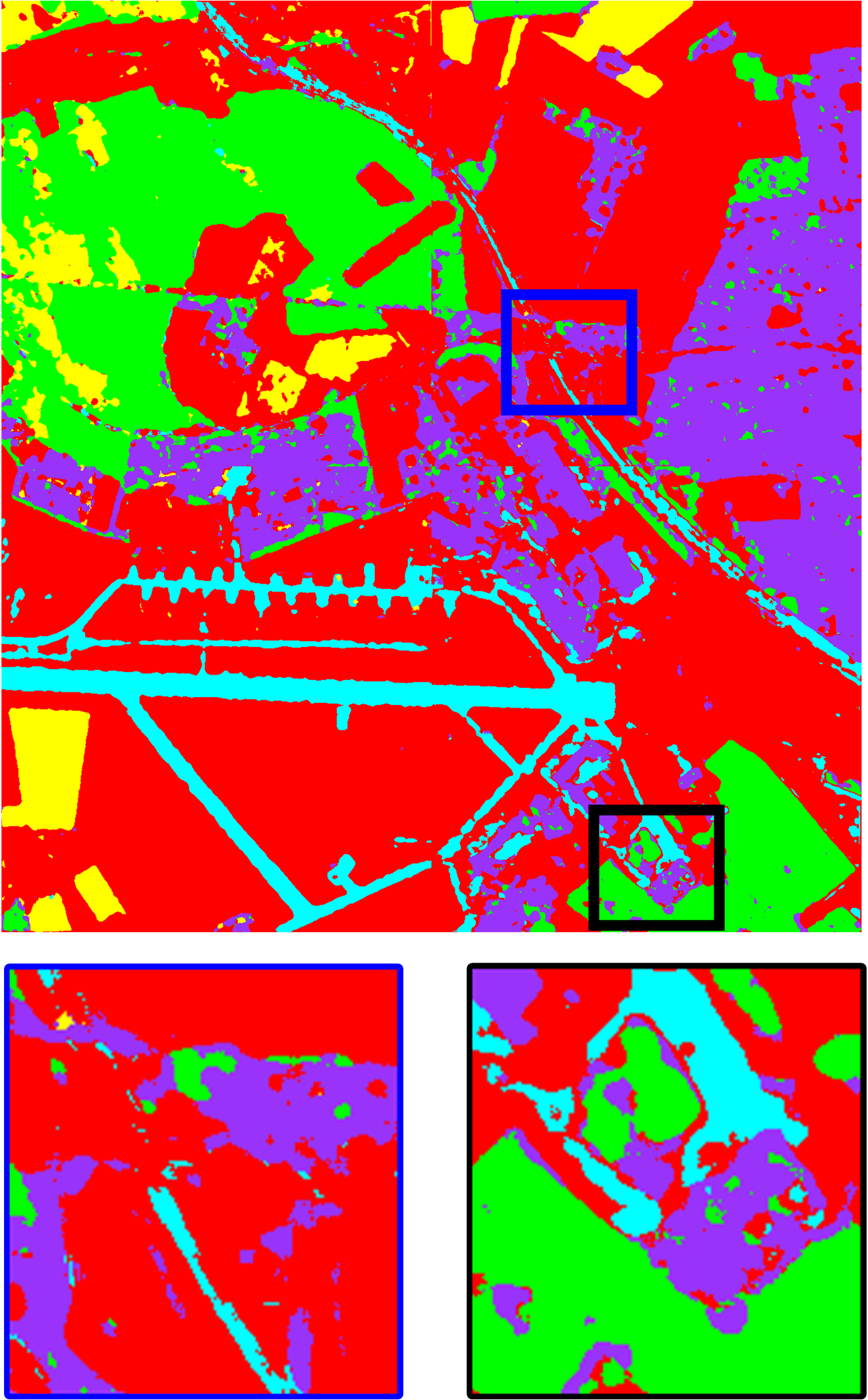}}
		\hspace{0.2mm}		\subfloat[]{\includegraphics[height=0.16\textheight]{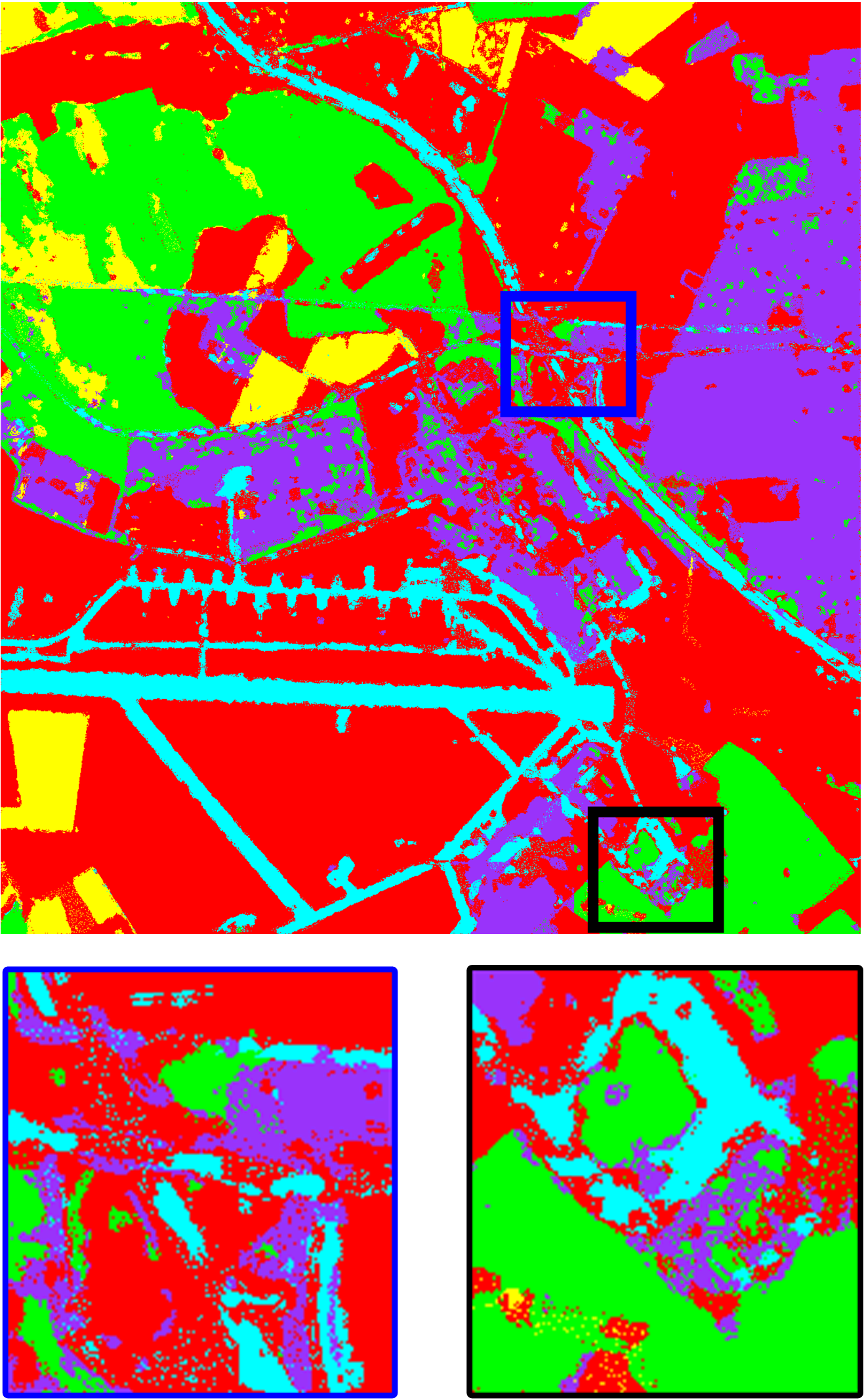}}
		\hspace{0.2mm}		\subfloat[]{\includegraphics[height=0.16\textheight]{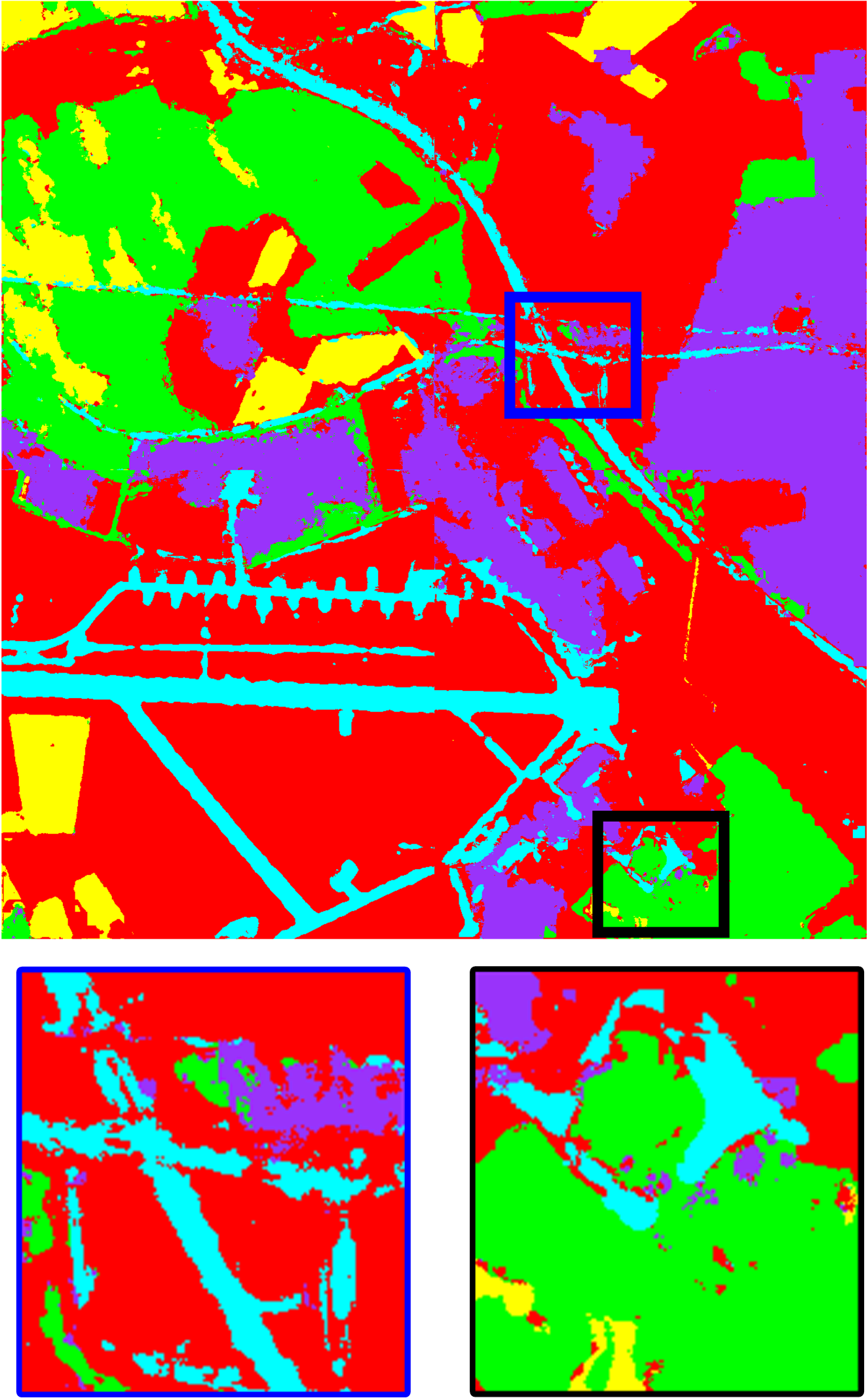}}
		\hspace{0.2mm}		\subfloat[]{\includegraphics[height=0.16\textheight]{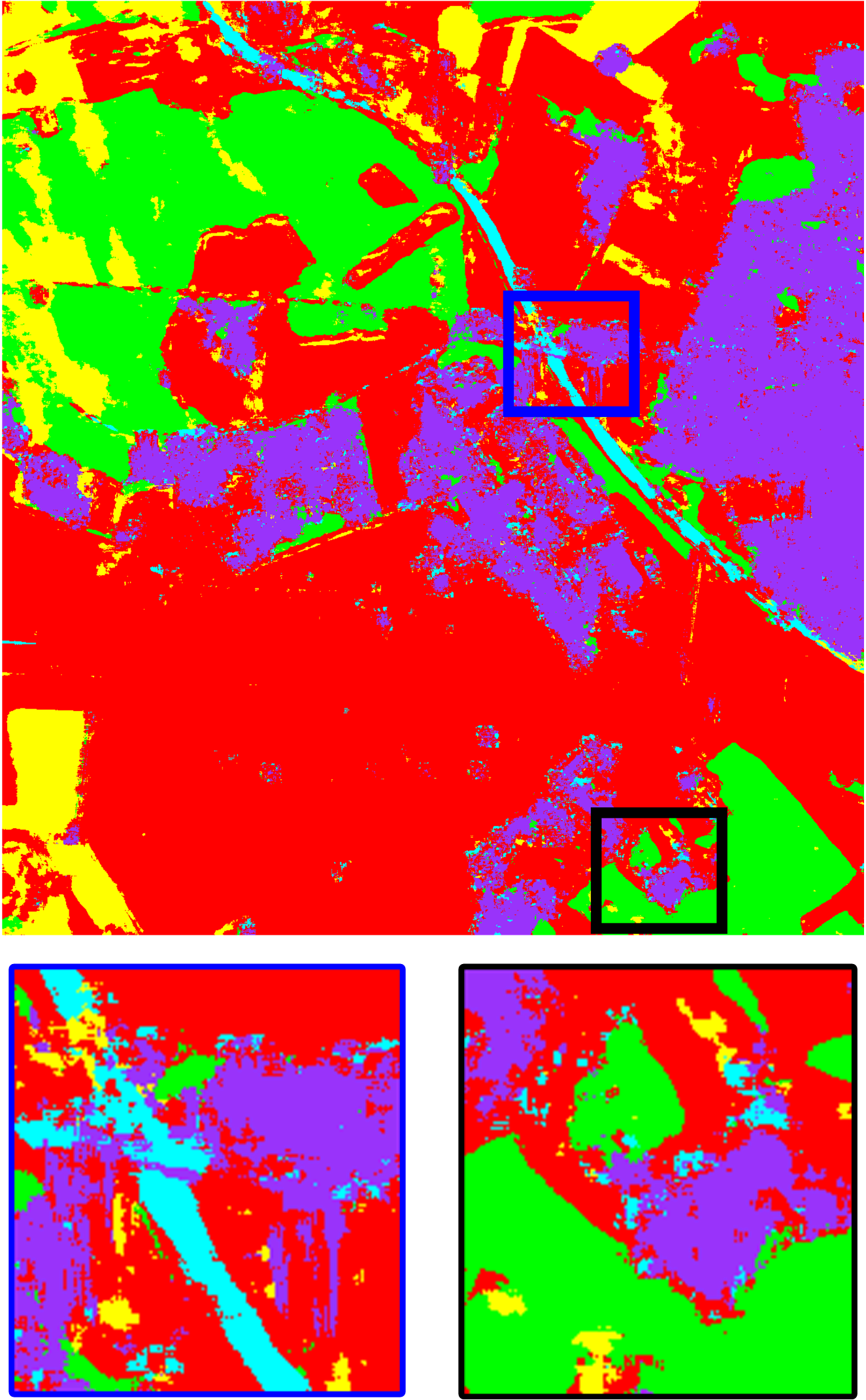}}
		\hspace{0.2mm}		\subfloat[]{\includegraphics[height=0.16\textheight]{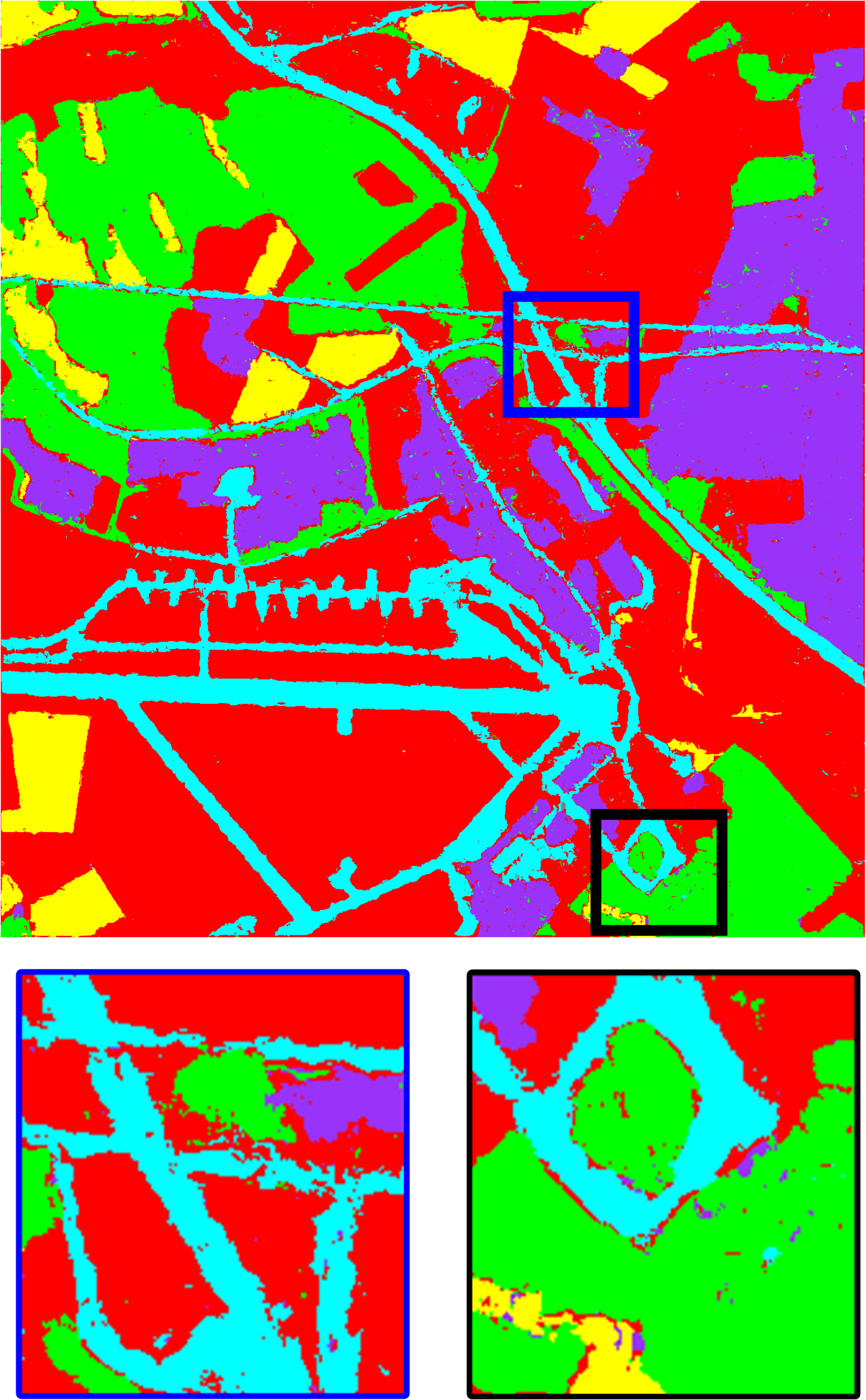}}
		
		
		\includegraphics[height=0.02\textheight]{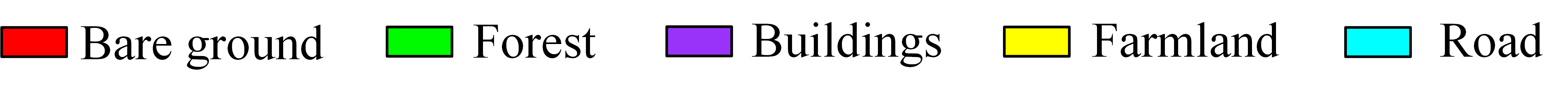}
		\caption{Classification results of the Oberpfaffenhofen data set. (a) KNNRS; (b) CVCNN; (c) 3DCNN; (d) DFGCN; (e) CEGCN; (f) PolMPCNN; (g) SRSR\_CNN. }
		\label{fig7}
	\end{figure*}

	Figures \ref{fig7}(a)-(g) display the classification results of the six compared methods and the proposed method, respectively. As depicted in Fig.\ref{fig7}(a), the KNNRS produces a lot of misclassified pixels similar to salt pepper noise due to the absence of polarization information. The CVCNN in (b) exhibits class confusion between \emph{bare ground} and \emph{farmland}. Additionally, some pixels in \emph{road} are misclassified as \emph{buildings} and \emph{bare ground}. Although the 3DCNN and the DFGCN are effective in reducing confusion, they lack the capability to classify heterogenous \emph{buildings} class. Compared to other methods, the CEGCN can achieve better classification performance in \emph{road}, which attributes to its utilization of pixel-wise local information and superpixel-wise spatial features. The PolMPCNN method nearly misclassifies the entire \emph{road} class due to its selection of a large patch size, which complicates the capturing of \emph{road} with complex structures. On the other hand, due to the lack of scattering features, this method results in misclassification of the boundary pixels, especially in \emph{road} and \emph{bare ground} classes. It is obvious that the proposed SRSR\_CNN has a superior classification effect, and the boundaries of \emph{road} is fine and smooth. This is because the Riemannian features learned from SRSRnet can enhance the classification performance at image boundaries by providing accurate and discriminative features.
	
Additionally, the classification accuracy of compared and proposed methods is presented in Table \ref{t2}. From the Table \ref{t2}, the proposed method demonstrates an excellent classification result, achieving higher OA of 5.9\%, 20.65\%, 12.71\%, 10.1\%, 6.6\%, and 19.71\% than other methods. The KNNRS in (a) obtains comparatively lower performance in \emph{farmland} and \emph{road}, with accuracies of 89.74\% and 75.82\%, respectively. The main misclassification in the CVCNN method is \emph{bare ground} and \emph{farmland} classes due to the lack of polarization information. The 3DCNN and DFGCN can not effectively classify the \emph{road} only using local information. The CEGCN shows low performance in \emph{farmland} and \emph{road}. The accuracy of \emph{road} in PolMPCNN is only 10.38\% due to the lack of matrix correlation. It is obvious that all the comparison methods can't classify the \emph{road} well, which indicates that the distinguishing features of this category cannot be extracted by using scattering features or local features alone. The proposed SRSR\_CNN integrates the matrices' geometric features and deep local features to enhance the classification accuracy.
	\vspace{-10pt}
	\begin{table*}[ht]
		\footnotesize
		\begin{center}
			\caption
			{ \label{t2}
				Classification accuracy of different methods on Oberpfaffenhofen Data Set(\%).}
			\begin{tabular}{p{2.2cm}|p{1.3cm}p{1.3cm}p{1.3cm}p{1.3cm}p{1.5cm}p{1.3cm}p{1.3cm}}
				\hline
				class&KNNRS&CVCNN&3DCNN&DFGCN&CEGCN&PolMPCNN&proposed\\
				\hline
				bare ground&91.33&68.86&91.76&90.51&93.14&86.92&\textbf{95.91}\\
				forest&92.26&81.16&84.59&85.08&90.13&82.51&\textbf{96.74}\\
				buildings&91.48&87.52&83.91&87.30&94.46&85.82&\textbf{95.98}\\
				farmland&89.74&70.49&65.33&77.57&80.70&65.65&\textbf{94.67}\\
				road&75.82&76.20&50.34&66.10&66.14&10.38&\textbf{91.96}\\
				\hline
				OA&89.63&74.88&82.82&85.43&88.93&75.82&\textbf{95.53}\\
				AA&88.12&76.85&75.19&81.31&84.91&66.26&\textbf{95.05}\\
				Kappa&84.93&65.73&74.28&78.66&83.66&63.49&\textbf{93.50}\\
				F1\_Score&87.69&71.31&77.94&82.37&86.44&64.48&\textbf{95.09}\\
				MIoU&78.47&56.32&64.63&70.32&76.64&52.05&\textbf{90.69}\\
				\hline
			\end{tabular}
		\end{center}
	\end{table*}

\subsection{Experimental results on Flevoland data set}

\begin{figure*}
	\centering
	\setlength{\fboxrule}{0.2pt}
	\setlength{\fboxsep}{0.01mm}
	\subfloat[]{\includegraphics[height=0.18\textheight]{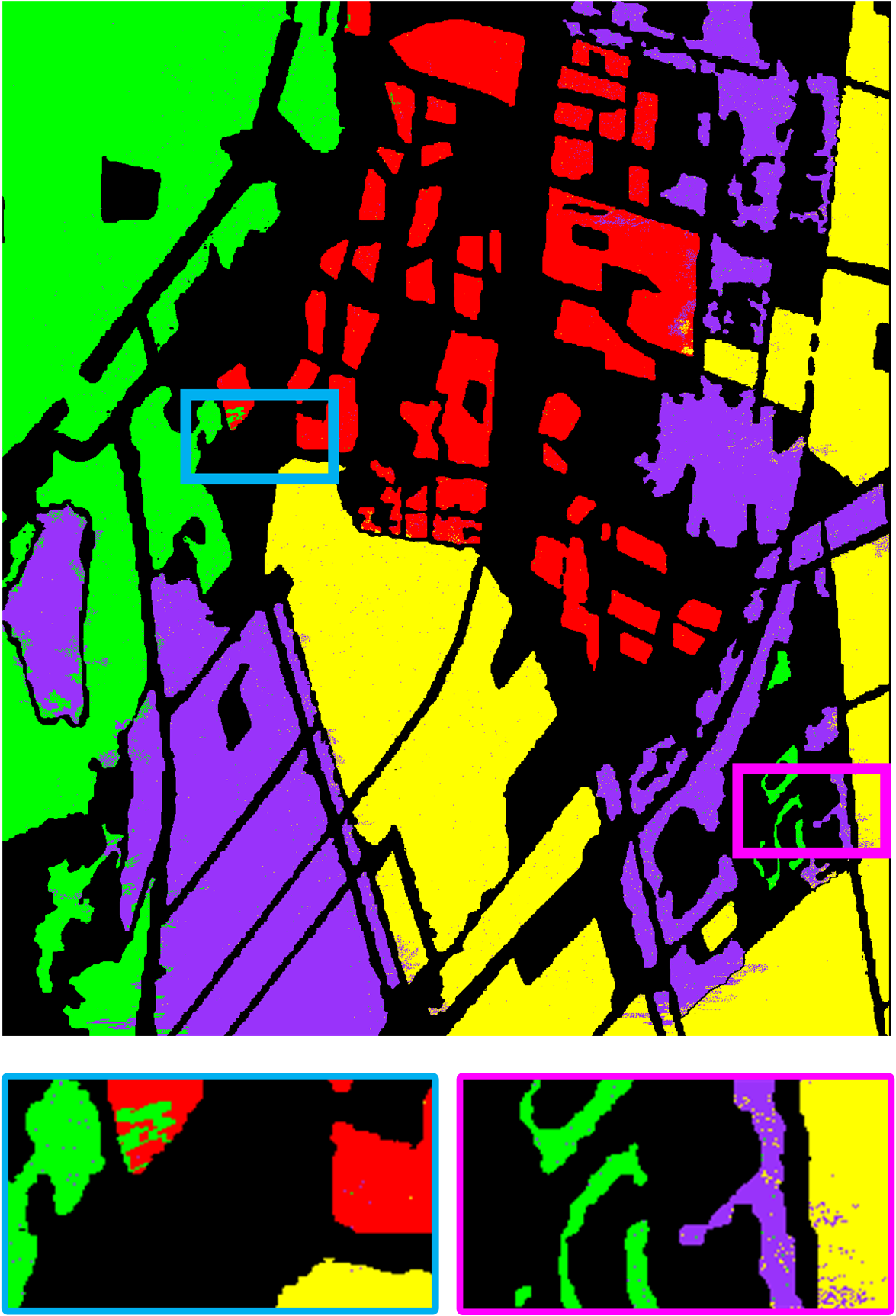}}
	\hspace{0.2mm}
	\subfloat[]{\includegraphics[height=0.18\textheight]{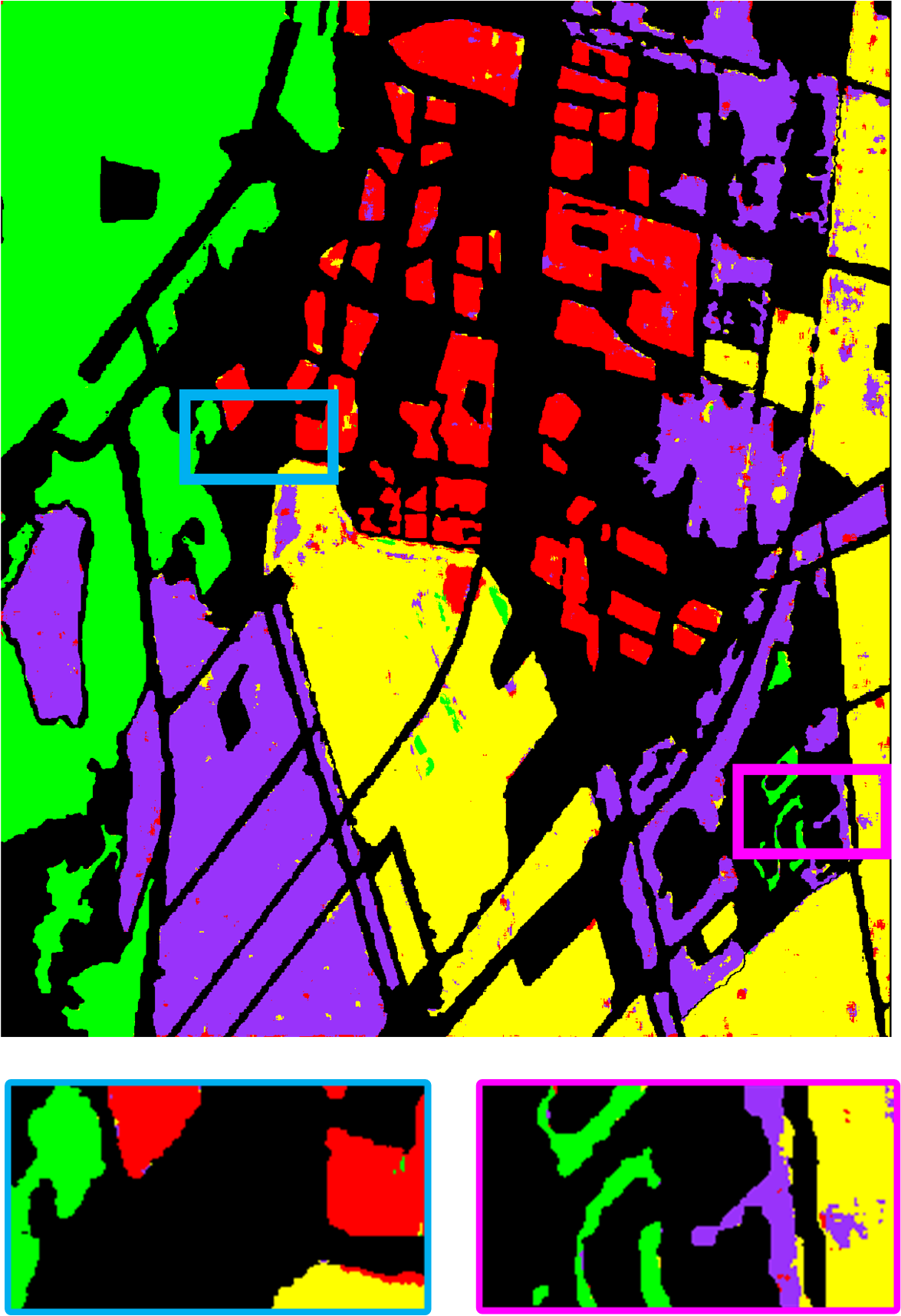}}
	\hspace{0.2mm}
	\subfloat[]{\includegraphics[height=0.18\textheight]{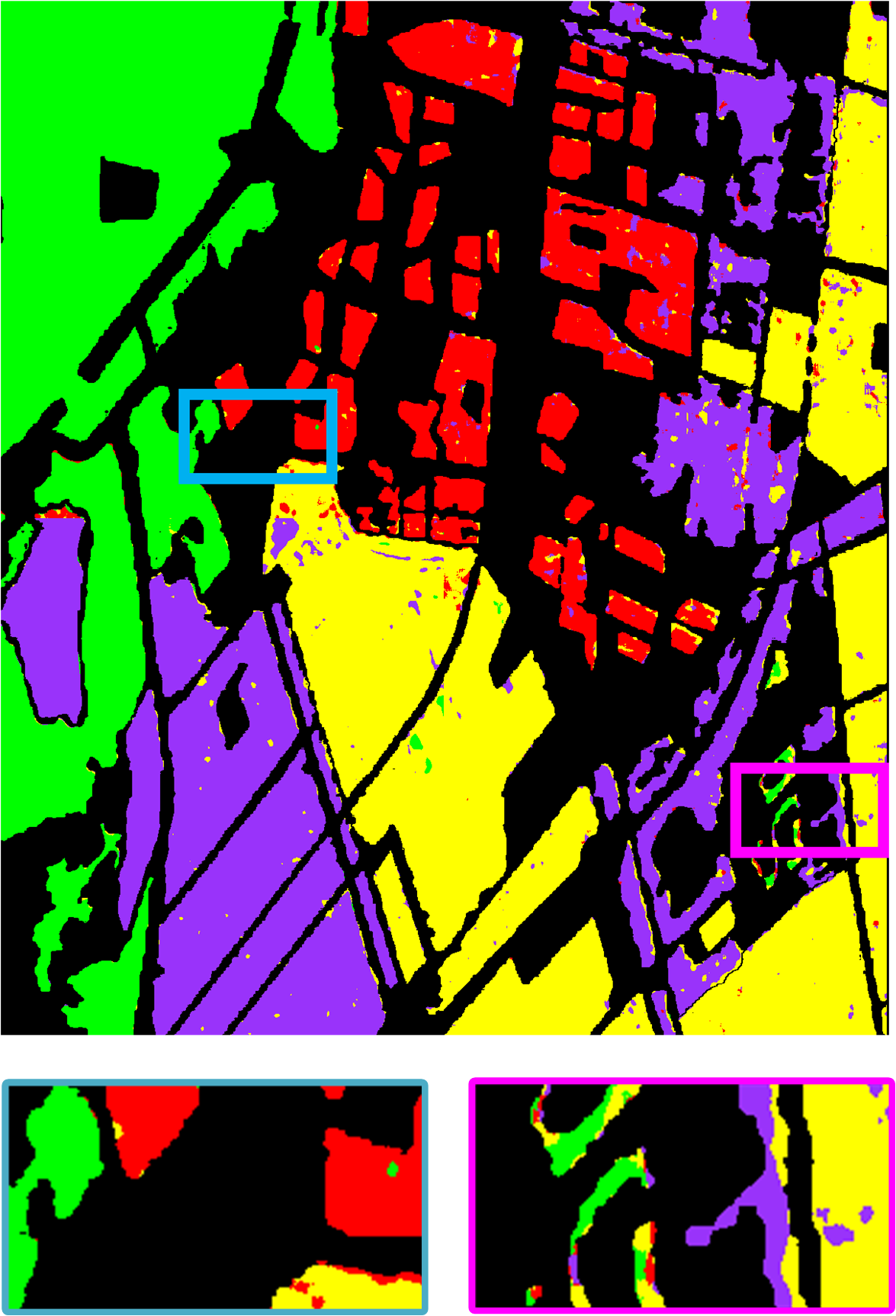}}
	\hspace{0.2mm}
	\subfloat[]{\includegraphics[height=0.18\textheight]{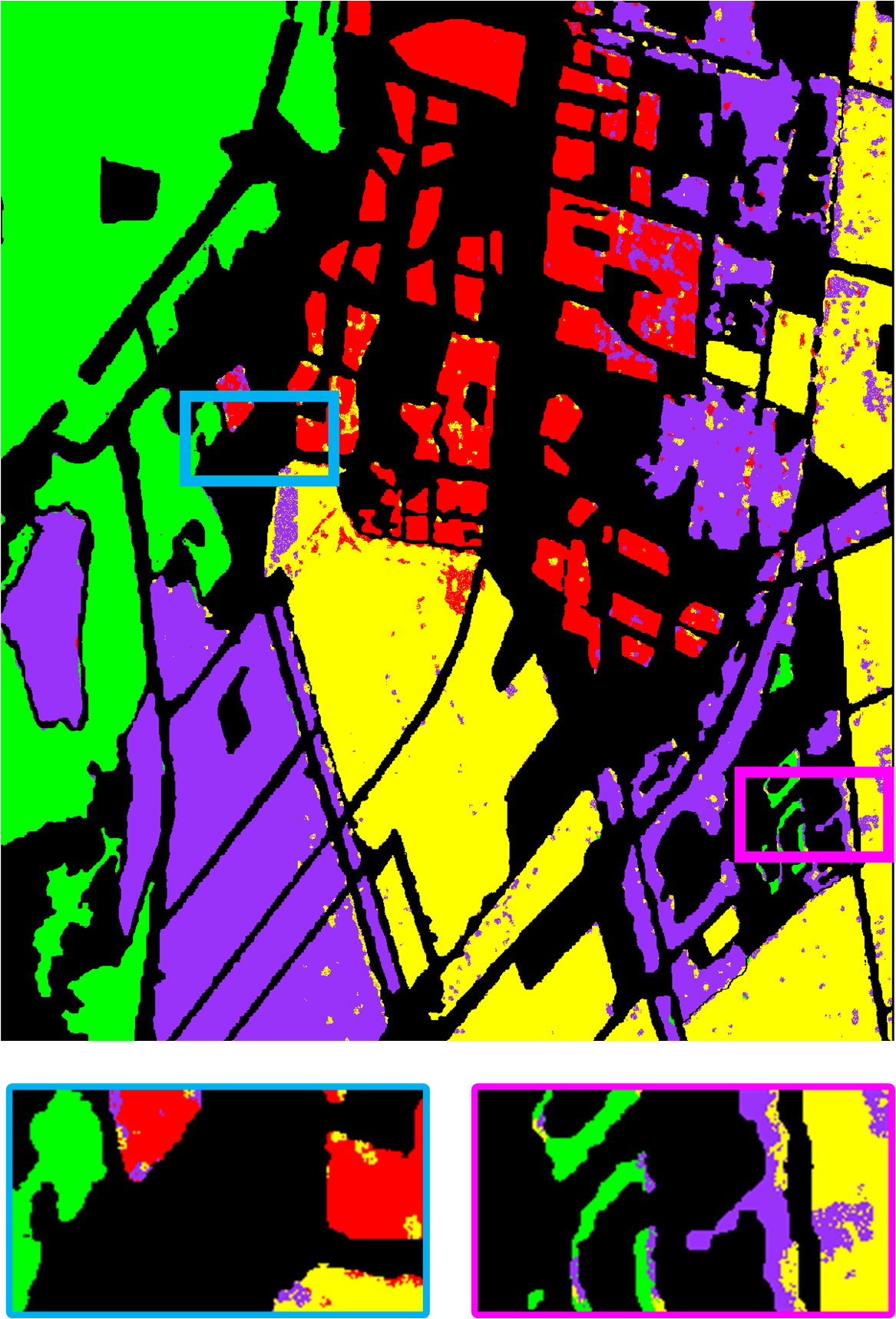}}
	\hspace{0.2mm}

	\subfloat[]{\includegraphics[height=0.18\textheight]{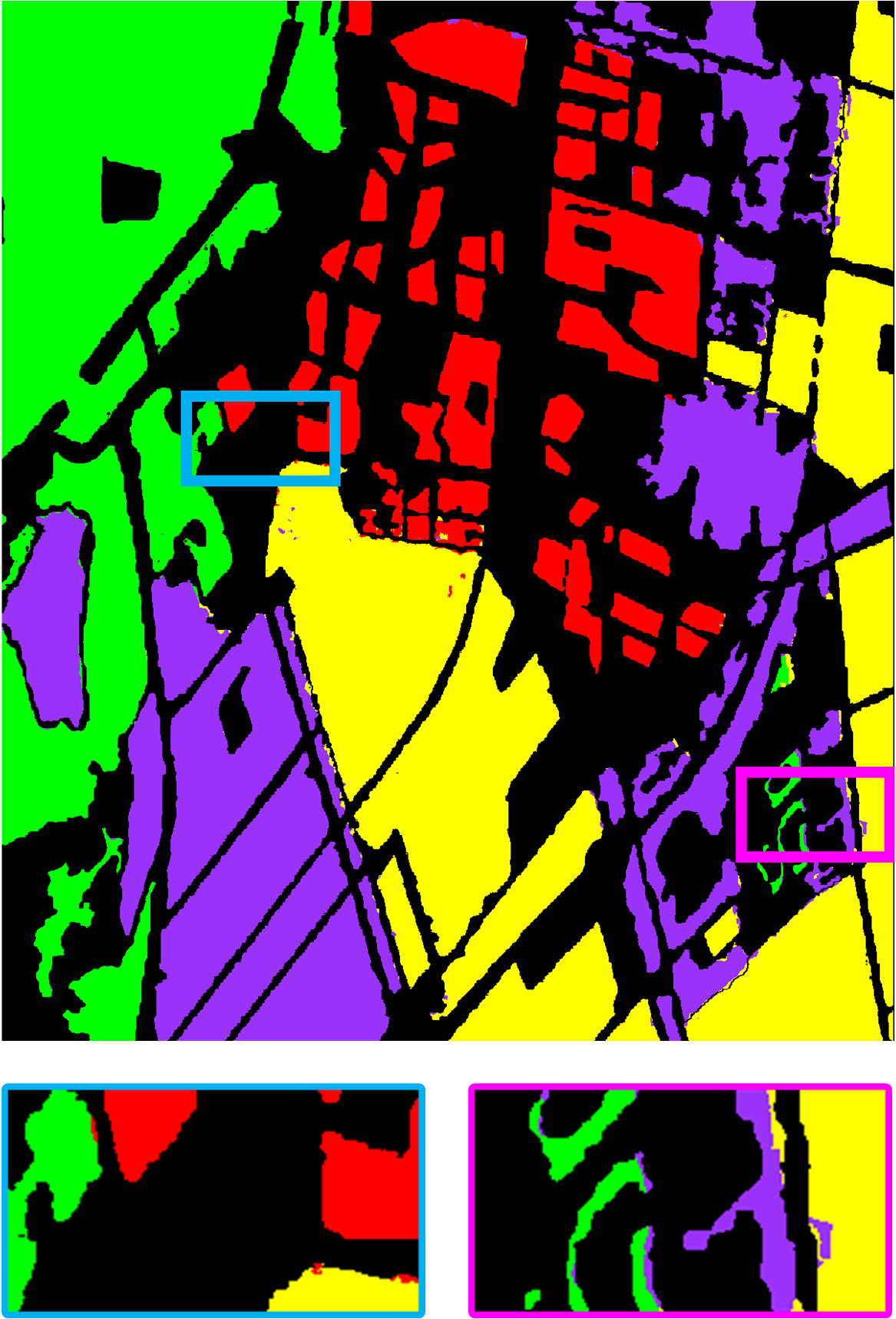}}
	\hspace{0.2mm}
	\subfloat[]{\includegraphics[height=0.18\textheight]{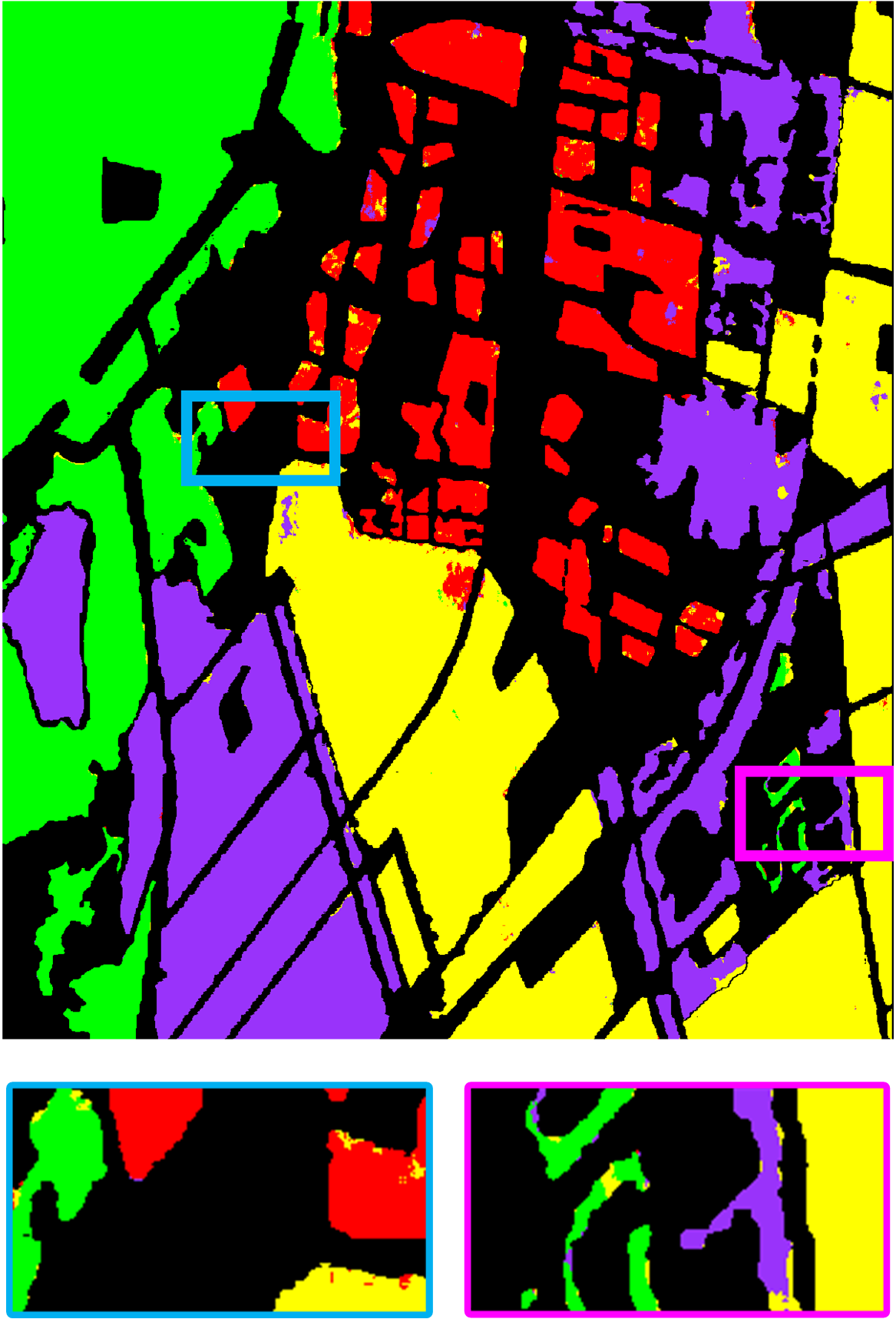}}
	\hspace{0.2mm}
	\subfloat[]{\includegraphics[height=0.18\textheight]{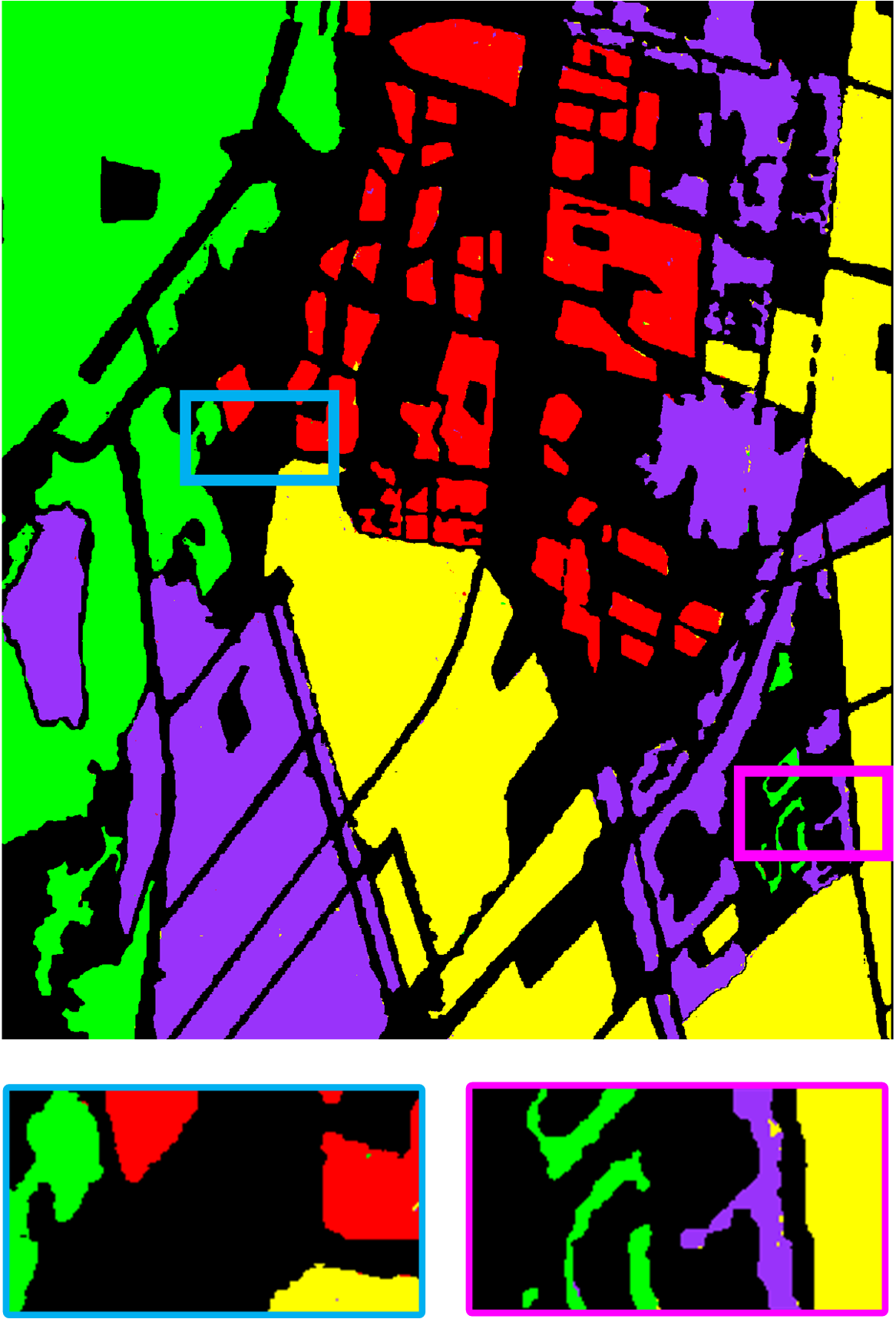}}
	
	
	\includegraphics[height=0.02\textheight]{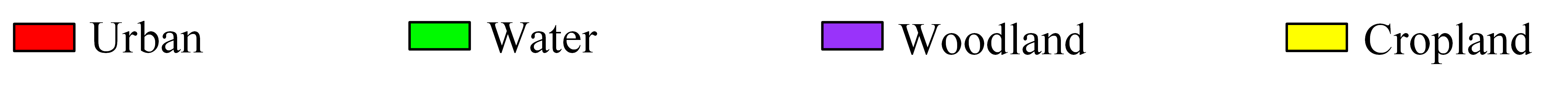}
	\caption{Classification results of the Flevoland data set. (a) KNNRS; (b) CVCNN; (c) 3DCNN; (d) DFGCN; (e) CEGCN; (f) PolMPCNN; (g) SRSR\_CNN. }
	\label{fig11}
\end{figure*}

Figures \ref{fig11}(a)-(g) illustrate the classification results of different methods. It is evident that all the methods have ideal classification preformance in \emph{water}. Specifically, the KNNRS in (a) produces some misclassifications in the boundary of \emph{water} and \emph{woodland}. The CVCNN can improve the result of edges by making fully use of scattering features, while there are still some misclassified pixels in \emph{urban}, \emph{woodland}, and \emph{cropland}. The 3DCNN, DFGCN and PolMPCNN also exhibit the confusion in distinguishing \emph{urban}, \emph{woodland}, and \emph{cropland}, since these methods are limited by their pixel-by-pixel feature learning approach. By utilizing 3DCNN and superpixel-wise GCN for extracting spatial information, the CEGCN can acquire distinctive features to enhance the classification results. However, some pixels on the boundary between \emph{woodland} and \emph{cropland} are mispredicted due to the lack of Riemannian features. The proposed SRSR\_CNN method effectively fuses the extracted matrix information and deep semantic features, resulting in improved classification result.

\begin{table*}[ht]
	\footnotesize
	\begin{center}
		\caption
		{ \label{t4}
			Classification accuracy of different methods on Flevoland Data Set(\%).}
		\begin{tabular}{p{2.2cm}|p{1.3cm}p{1.3cm}p{1.3cm}p{1.3cm}p{1.5cm}p{1.3cm}p{1.3cm}}
			\hline
			class&KNNRS&CVCNN&3DCNN&DFGCN&CEGCN&PolMPCNN&proposed\\
			\hline
			urban&98.03&96.26&94.74&91.83&\textbf{99.45}&96.29&99.40\\
			water&99.02&99.85&98.87&99.57&99.65&99.14&\textbf{99.94}\\
			woodland&97.73&96.48&96.03&95.93&98.96&98.77&\textbf{99.57}\\
			cropland&99.23&93.94&96.75&93.82&99.47&98.66&\textbf{99.86}\\
			\hline
			OA&98.59&96.57&96.84&95.69&99.37&98.49&\textbf{99.74}\\
			AA&98.50&96.63&96.60&95.29&99.38&98.21&\textbf{99.69}\\
			Kappa&98.07&95.33&95.69&94.11&99.14&97.94&\textbf{99.64}\\
			F1\_Score&98.62&96.31&96.72&95.45&99.40&98.31&\textbf{99.73}\\
			MIoU&97.29&92.95&93.70&91.41&98.81&96.70&\textbf{99.45}\\
			\hline
		\end{tabular}
	\end{center}
\end{table*}

Furthermore, to quantitatively assess the classification performance, the classification accuracies of different methods are shown in Table \ref{t4}. As indicated in Table \ref{t4}, the proposed method can obtain the highest classification accuracy under all evaluation indicators. Specifically, the proposed SRSR\_CNN achieves 1.15\%, 3.17\%, 2.9\%, 4.05\%, 0.37\%, and 1.25\% higher in OA than other comparative methods, respectively. The main misclassification of KNNRS method is the \emph{urban} class, primarily due to the absence of scattering characteristics. Additionally, the CVCNN exhibits relatively lower accuracy in the \emph{cropland} class, indicating that pixel-wise methods struggle to capture the features of heterogeneous regions. The 3DCNN, DFGCN, and PolMPCNN also suffer from obvious false classification in the \emph{urban} class due to the strength of heterogeneous features. However, the CEGCN achieves the best classification performance in \emph{urban} areas by utilizing superpixel-wise GCN to learn discriminative features. The proposed SRSR\_CNN can attain superior classification performance by capturing high-level features.

\subsection{Ablation Study}

There are three modules comprising of the proposed method, including SRSR, SRSRnet and CNN-enhanced modules. In addition, the SRSRnet module consists of sparse coefficient learning(SCL) and dictionary learning(DL) in each layer. Here, we conduct ablation experiments on the aforementioned three datasets to further validate the contributions of different modules of the proposed SRSR\_CNN. In addition, to verify each part in the SRSRnet, the SRSRnet with only SCL by fixing dictionary atoms is utilized as the ablation study. Evaluation metrics include OA and Kappa , and the classification results are shown in Table \ref{t5}. To be specific, with only CNN module, we utilize the original 9-dimension feature as the input of CNN, which is obtained from vectoring the covariance matrix. The SRSR module without unfolding network employs the SPG algorithm to obtain the sparsity coefficients. During the SRSRnet module, if the dictionary atoms are fixed, we only learn sparse coefficient in each layer, which verifies the effectiveness of dictionary learning.

It is obvious from Table \ref{t5} that the classification accuracy with only the CNN in Euclidean space is low. The classification accuracy after adding the SRSR module has a significant improvement, which verifies the effectiveness of SRSR model. The SRSRnet can gradually improves classification accuracy by adding sparse coefficient and dictionary atom learning. The proposed SRSR\_CNN method can combine the feature learning abilities of both SRSRnet and CNN modules to improve classification performance.

%

\begin{table}[ht]
	\footnotesize
	\begin{center}
		\caption
		{ \label{t5}
			Classification accuracy of different modules on three Data sets(\%).}
		\begin{tabular}{p{1.4cm}|p{0.6cm}p{0.6cm}p{0.6cm}p{0.6cm}|p{0.6cm}p{0.6cm}}
			\hline
			Dataset&SRSR&SCL&DL&CNN&OA&Kappa\\
			\hline
			\multirow{4}{*}{Xi'an}&\ding{55}&\ding{55}&\ding{55}&\ding{51}&91.18&85.61\\
			&\ding{51}&\ding{55}&\ding{55}&\ding{51}&93.13&88.69\\
			&\ding{51}&\ding{51}&\ding{55}&\ding{51}&96.80&94.72\\
			&\ding{51}&\ding{51}&\ding{51}&\ding{51}&96.98&95.01\\
			\hline
			\multirow{4}{*}{Oberpfaffenhofen}&\ding{55}&\ding{55}&\ding{55}&\ding{51}&83.90&76.01\\
			&\ding{51}&\ding{55}&\ding{55}&\ding{51}&89.68&84.98\\
			&\ding{51}&\ding{51}&\ding{55}&\ding{51}&95.03&92.81\\
			&\ding{51}&\ding{51}&\ding{51}&\ding{51}&95.53&93.50\\
			\hline
			\multirow{4}{*}{Flevoland}&\ding{55}&\ding{55}&\ding{55}&\ding{51}&97.48&96.56\\
			&\ding{51}&\ding{55}&\ding{55}&\ding{51}&98.88&98.47\\
			&\ding{51}&\ding{51}&\ding{55}&\ding{51}&99.38&99.16\\
			&\ding{51}&\ding{51}&\ding{51}&\ding{51}&99.74&99.64\\
			\hline
		\end{tabular}
	\end{center}
\end{table}

\subsection{Effect of the superpixel scale parameter $\delta $}

\begin{figure*}
	\centering
	\setlength{\fboxrule}{0.2pt}
	\setlength{\fboxsep}{0.01mm}
    \subfloat[]{\includegraphics[height=0.16\textheight]{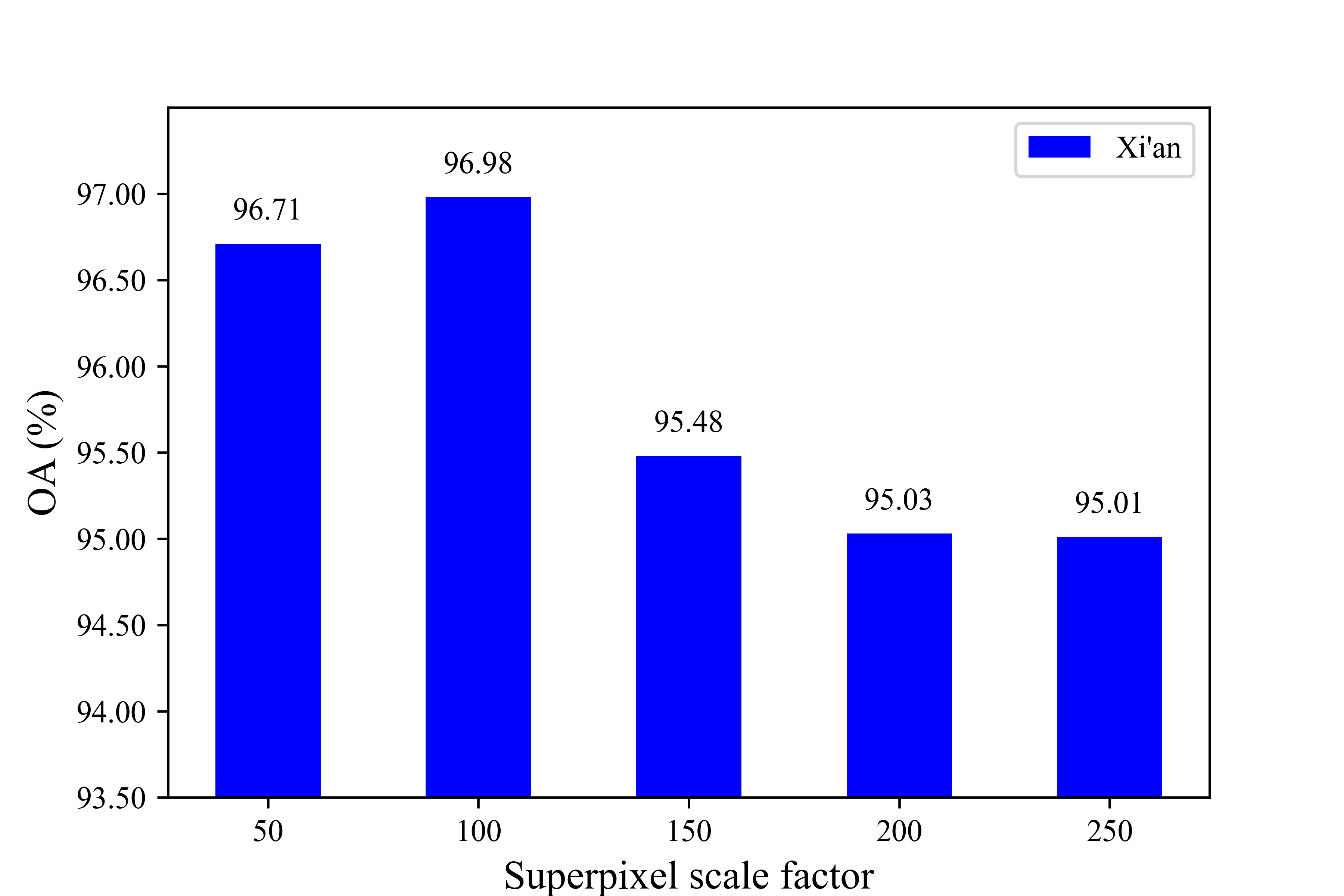}}
	\subfloat[]{\includegraphics[height=0.16\textheight]{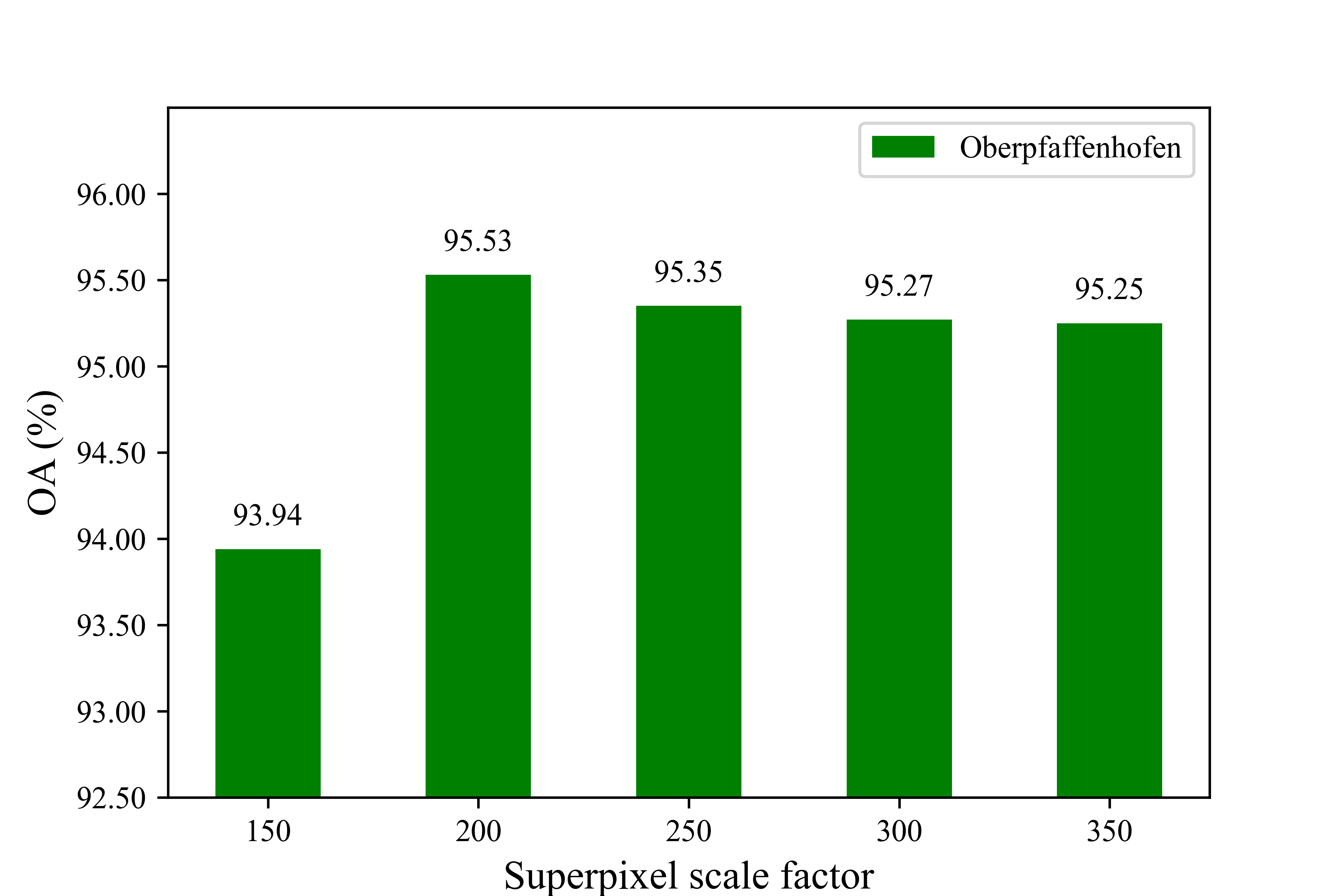}}
	\subfloat[]{\includegraphics[height=0.16\textheight]{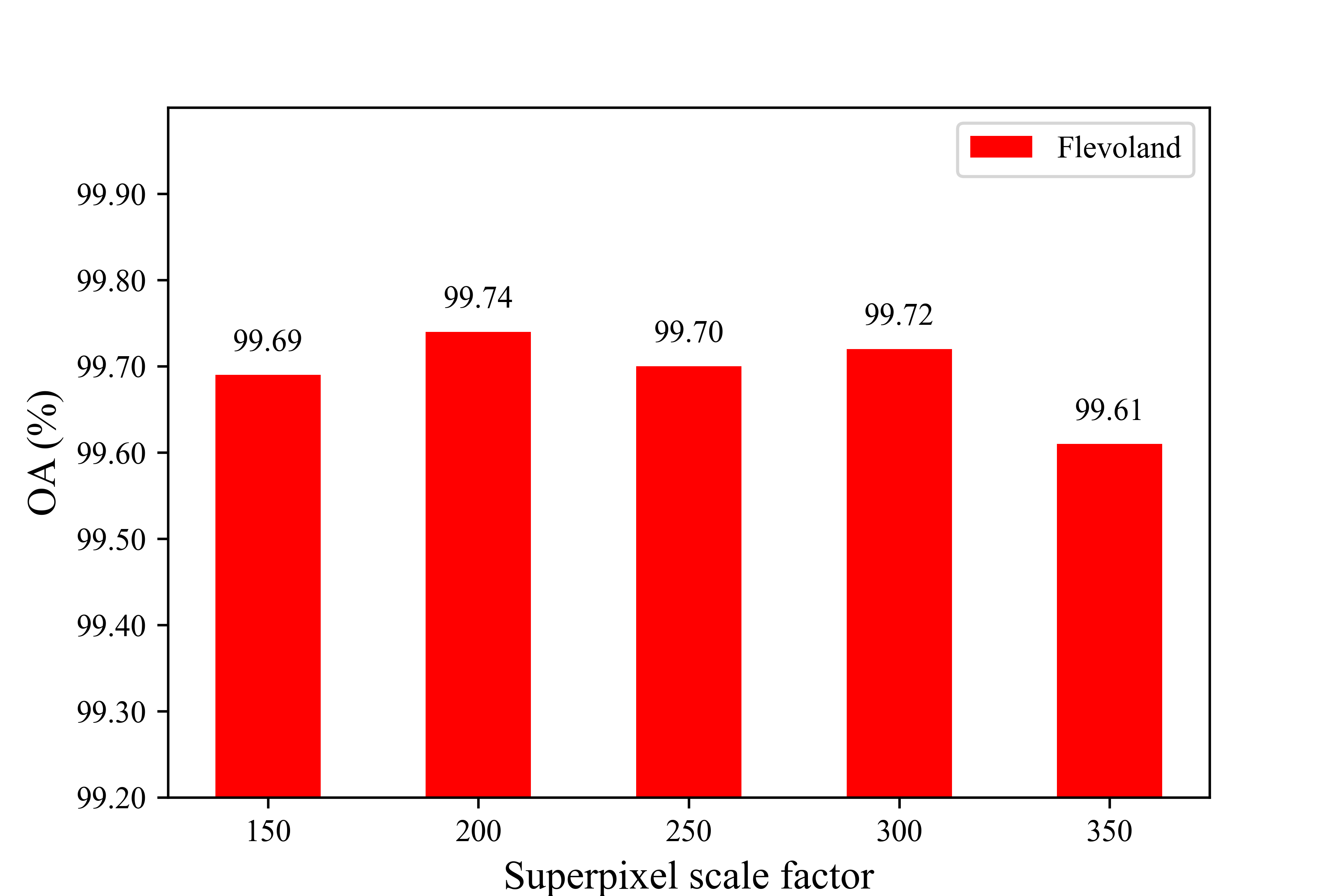}}
	\caption{The effect of superpixel scale parameter on classification accuracy. (a) Xi'an dataset. (b) Oberpfaffenhofen dataset. (c) Flevoland dataset.}
	\label{fig13}
\end{figure*}

The superpixel scale parameter $\delta $ is an essential parameter for superpixel segmentation, which represents the average superpixel size. A smaller $\delta $ means more superpixels which apparently increase the computing cost, while too large $\delta $ will have less superpixels which causes the edge confusion. Therefore, it is crucial to select an appropriate superpixel scale factor. Here, we conducted experiments using various superpixel scales on three datasets. The impact of the superpixel scale parameter on classification accuracy is depicted in Figs.\ref{fig13}(a)-(c). Figure \ref{fig13}(a) illustrates the OA achieved by our proposed SRSR\_CNN on the Xi'an dataset, utilizing superpixel scales ranging from 50 to 250. In addition, for large-scale PolSAR images, we vary the scales from 150 to 350 on the Oberpfaffenhofen and Flevoland datasets, as shown in Figs.\ref{fig13}(b)-(c). It can be seen that the classification accuracy reaches the peak and changes relatively smoothly when $\delta  = 200$ for large-scale images. For small-scale images, the classification effect is the best when $\delta  = 100$. Therefore, We can select the scale parameter in range of 200 to 250 for large-scale images, and choose 100 for the small-scale images.

\vspace{-10pt}

\begin{figure*}
	\centering
	\setlength{\fboxrule}{0.2pt}
	\setlength{\fboxsep}{0.01mm}
	\subfloat[]{\includegraphics[height=0.16\textheight]{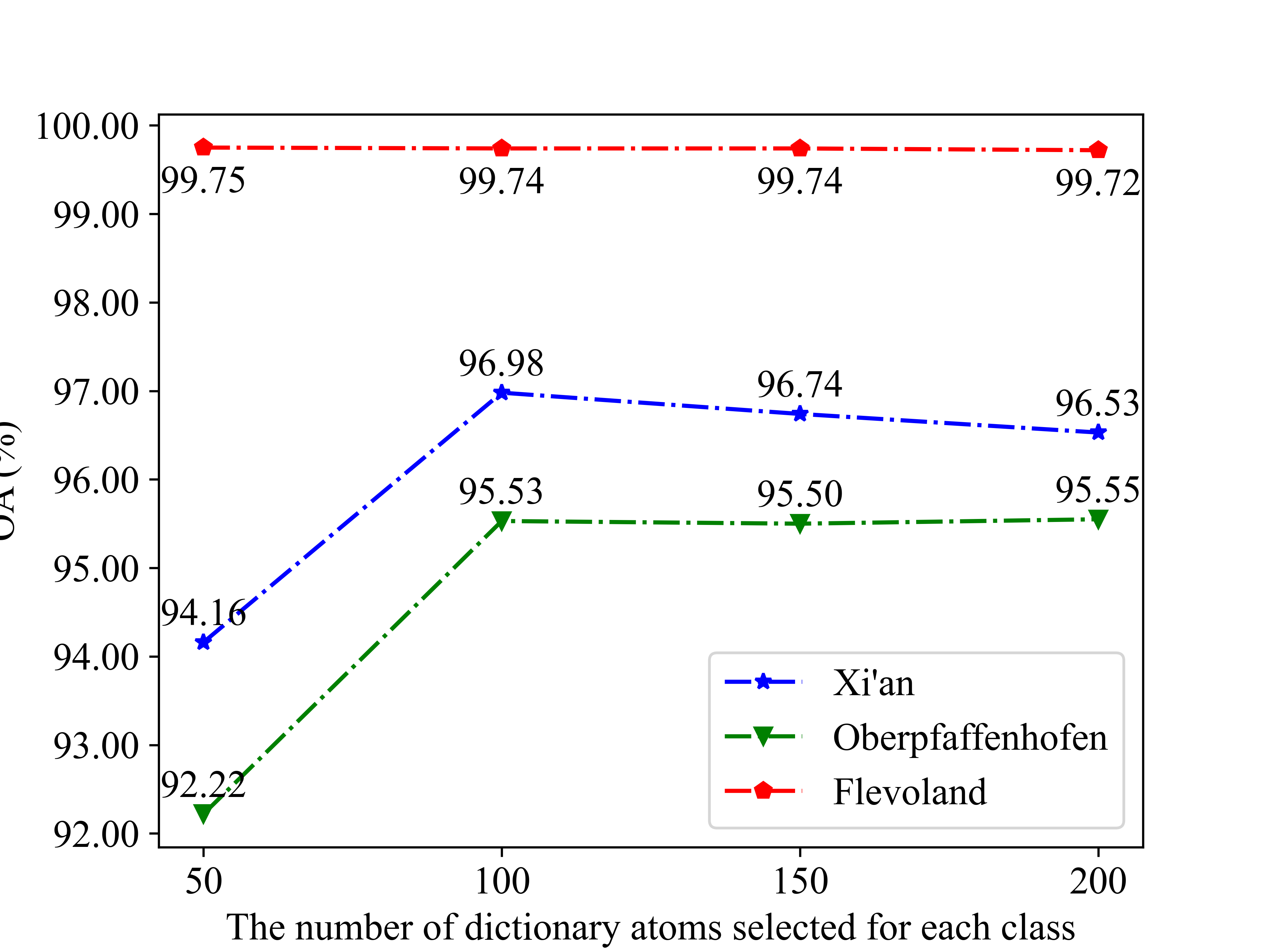}}
	\subfloat[]{\includegraphics[height=0.16\textheight]{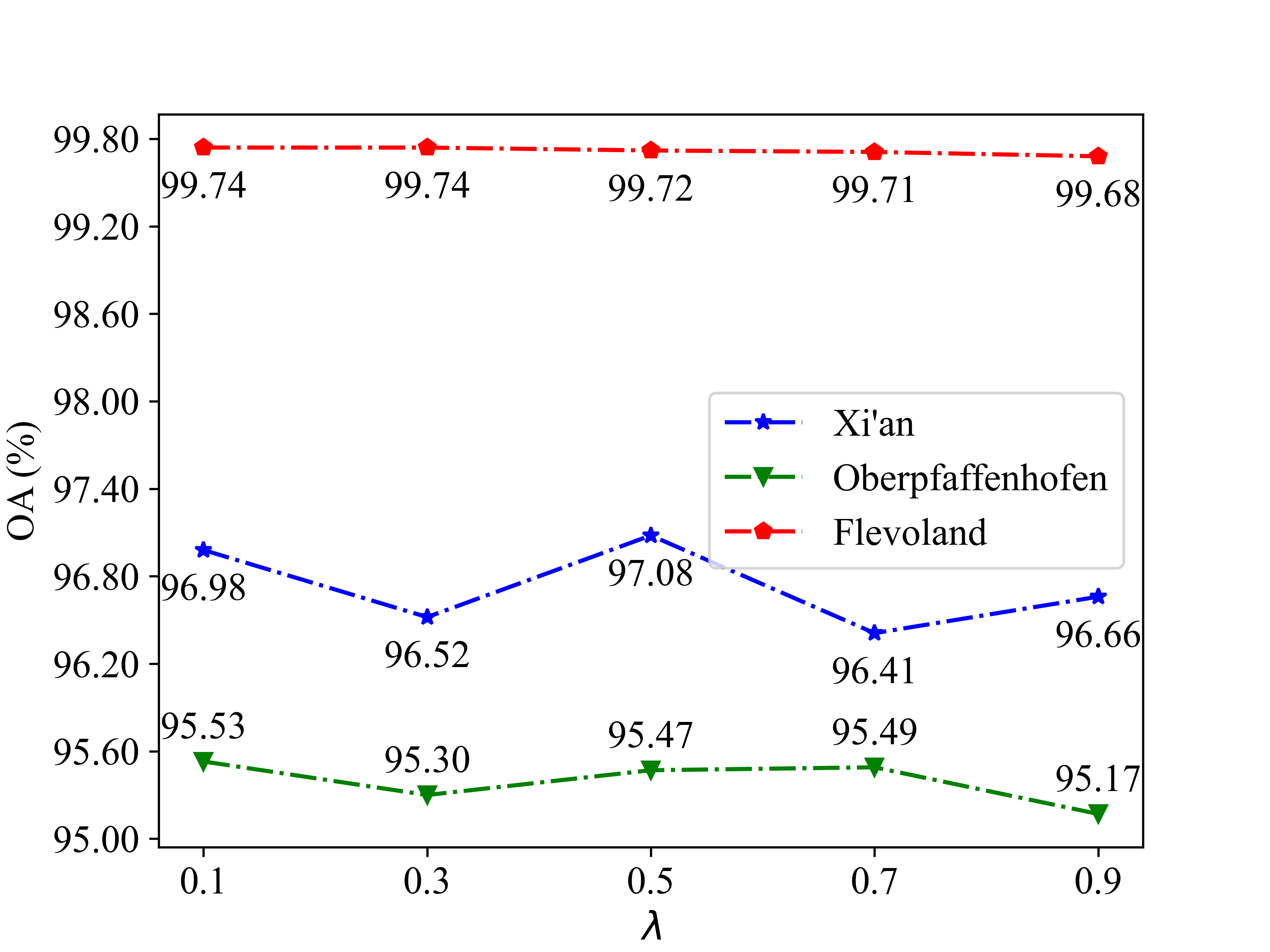}}
	\subfloat[]{\includegraphics[height=0.16\textheight]{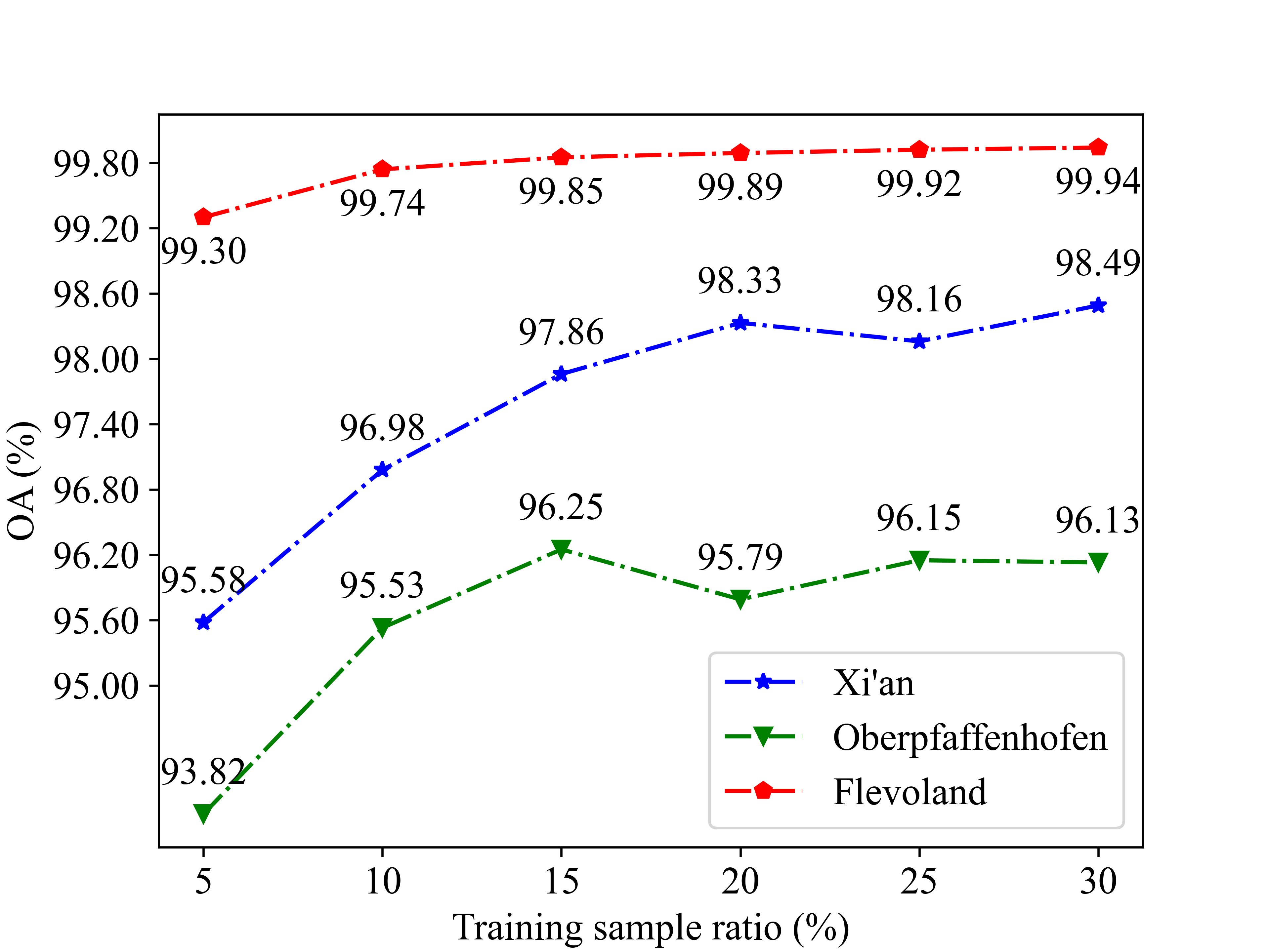}}
	\caption{(a)The effect of the number of dictionary atom; (b)Effect of the regularization parameter $\lambda $; (c)The effect of the ratio of training samples on classification accuracy.}
	\label{fig16}
\end{figure*}

\subsection{Effect of the number of dictionary atom}

The number of dictionary atom is the essential parameter for sparse representation. In this paper, we test the effect of dictionary size on classification performance by ranging dictionary size from 50 to 200 with an interval of 50. Figure \ref{fig16}(a) shows the variation of classification accuracy with different dictionary sizes on three PolSAR datasets. As depicted in Fig.\ref{fig16}(a), the Flevoland dataset exhibits minor fluctuations in OA. When the number of selected dictionary atoms falls within the range of 50 to 100, there is a significant improvement in OA for the Xi'an and Oberpfaffenhofen datasets. What's more, when the number of atoms reaches 100, the growth trend of OA remains relatively constant. However, as the number of atoms increases, more matrix operations are involved, leading to a substantial rise in computation. Therefore, we select 100 dictionary atoms for each class to balance the computational cost and classification accuracy.

\subsection{Effect of the regularization parameter $\lambda $}

The regularization parameter $\lambda $ is another important parameter of the Riemannian sparse representation model, which controls the contribution of sparse term. Its typical set is within the range of 0 to 1. In this experiment, we investigate the impact of varying values on OA, as shown in Fig.\ref{fig16}(b). To be specific, we vary it from 0.1 to 0.9 with the interval of 0.2. According to the Fig.\ref{fig16}(b), it indicates that the OA remains relatively stable for the Flevoland dataset. However, for the Xi'an and Oberpfaffenhofen datasets, there is a fluctuation range of about 0.5\%. Additionally, we can see that the peak accuracy exists around 0.1 for four datasets. Therefore, the $\lambda $ is set to 0.1 in this paper.

\subsection{Effect of the training sample ratio}

The training sample ratio plays a crucial role in performance of network learning. To select the optimal training sample ratio, we discuss the changes of OA on three datasets with different training sample ratios in Fig.\ref{fig16}(c). It is clear that the training accuracy improves significantly, as the training sample ratio increases from 5\% to 10\%. However, when the training sample ratio reaches 10\%, the trend becomes relatively stable, and OA shows no significant improvement. In addition, too many training samples will cause unbearable computation cost. Therefore, we select 10\% as the training sample ratio in this experiment.

\subsection{Analysis of running time}

We utilize the Xi'an dataset as a case study to analyze the running time of different methods. Table \ref{t6} presents the training and testing times of the compared and the proposed SRSR\_CNN methods. In particular, the PolMPCNN exhibits the longest training and test times which can be attributed to its input feature dimension and large-scale convolution. On the other hand, the 3DCNN demonstrates the shortest training time due to its 3D calculation. The running time of CEGCN is also shorter, since it uses superpixel-level GCN to extract features. The proposed method can obtain the best classification performance within a relatively short time. It demonstrates the effectiveness of the proposed method in terms of both time efficiency and performance.
\begin{table}[ht]
	\footnotesize
	\begin{center}
		\caption
		{ \label{t6}
			Computing time of different methods on Xi'an Data Set ($s$)}
		\begin{tabular}{p{0.6cm}p{0.6cm}p{0.6cm}p{0.6cm}p{0.6cm}p{0.6cm}p{1cm}p{0.6cm}}
			\hline
			&KNNRS&CVCNN&3DCNN&DFGCN&CEGCN&PolMPCNN&proposed\\
			\hline
			training&4718.59&463.20&121.84&475.62&129.43&26100.35&297.93\\
			testing&26.21&38.43&22.80&7.85&4.45&327.53&13.69\\
			\hline
		\end{tabular}
	\end{center}
\end{table}

\vspace{-4pt}
\subsection{Feature Visualizations by t-SNE}

\begin{figure}
	\centering
	\setlength{\fboxrule}{0.2pt}
	\setlength{\fboxsep}{0.01mm}
	\includegraphics[height=0.18\textheight]{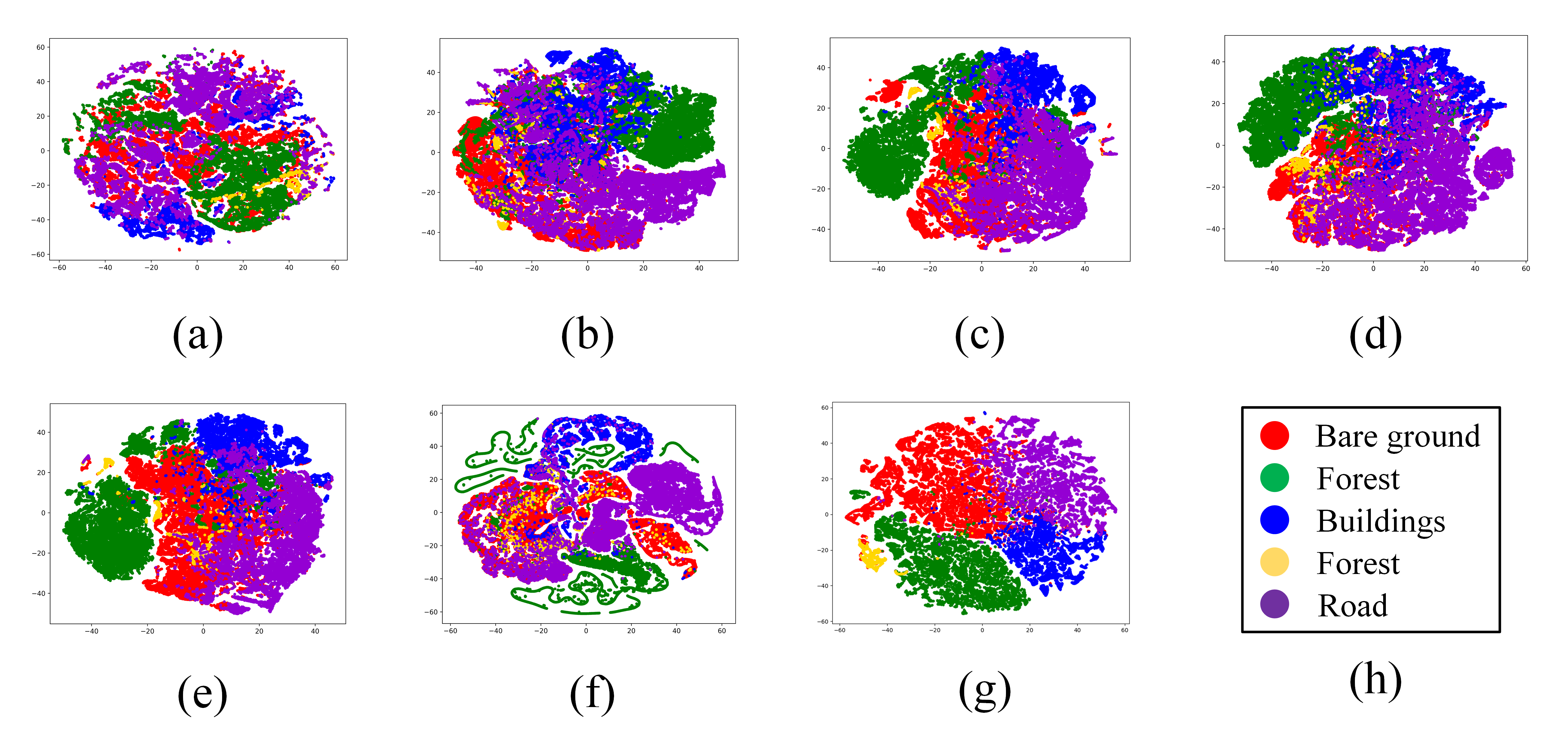}
	\caption{t-SNE visualization results of different methods on Oberpfaffenhofen data set. (a) KNNRS; (b) CVCNN; (c) 3DCNN; (d) DFGCN; (e) CEGCN; (f) PolMPCNN; (g) proposed; (h) label map.}
	\label{fig17}
\end{figure}

In order to evaluate the feature representation ability of the proposed SRSR\_CNN more intuitively, we use $t$-distributed stochastic neighbor embedding ($t$-SNE)\cite{tsne} to visualize the feature distribution of the above comparison methods and the proposed method. The learned features by different methods are visualized on Oberpfaffenhofen data set, and the visualization results are shown in Fig. \ref{fig17}. We can see that the proposed SRSR\_CNN demonstrates better separability, resulting in only a small number of misclassified samples. Moreover, the boundaries between clusters are well-defined, demonstrating the robustness of the proposed method in distinguishing among different classes.

\vspace{-10pt}
\section{Conclusion}

This paper presents a novel Riemannian sparse representation learning network (SRSR\_CNN) for PolSAR image classification.  It is the pioneer work to construct a sparse representation-guided deep learning network for PolSAR images. The proposed method consists of three modules: superpixel-based Riemannian sparse representation (SRSR) model,  SRSRnet module, and CNN-enhanced module. Accordingly, the sparse model-guided learning network is developed for PolSAR image classification for the first time.  Firstly, an SRSR model is constructed by defining covariance matrix dictionary sets. Then, the optimization method is inferred, and the optimization procedure is unfolded into an SRSR network structure to learn sparse coefficients and dictionry. Finally, a CNN-enhanced module is designed to learn high-level features, improving the classification performance. Experiments demonstrate the proposed method can achieve excellently quantitative and qualitative results in region homogeneity and edge preservation accuracy compared to state-of-the-art methods.

\section*{Acknowledgments}
This work was supported in part by the National Natural Science Foundation of China under Grant 62006186,62272383, the Youth Innovation Team Research Program Project of Education Department in Shaanxi Province under Grant 23JP111.

{\appendix[Riemannian CG algorithm]
As shown in the section III, the dictionary update subproblem can be described by the formula as:
\begin{equation}\label{24}
	\begin{gathered}
		{D^*} = \min \theta (D) = \frac{1}{2}\left\| {\log (\sum\limits_{i = 1}^N {{\alpha _i}} {X_{\overline S }}^{ - \tfrac{1}{2}}{D_i}{X_{\overline S }}^{ - \tfrac{1}{2}})} \right\|_F^2 \\
		+ {\lambda _B}\sum\limits_{i = 1}^N {Tr\left( {{D_i}} \right)}  \\
	\end{gathered}
\end{equation}

The model in Equ.(\ref{24}) is a non-convex optimization problem, for which an analytical resolution is not available. Conjugate gradient (CG)\cite{RN150}, also known as conjugate gradient descent, is a classic iterative optimization algorithm that can be used to solve specific unconstrained optimization problems. The Riemannian CG algorithm is to extend the CG to the Riemannian space $M$. The specific solution process is as follows. For a smooth nonlinear function $\begin{array}{*{20}{c}}
	{\psi \left( x \right),}&{x \in {R^n}}
\end{array}$, the iterative formula of the CG method can be expressed as:
\begin{equation}\label{25}
	{x^{k + 1}} = {x^k} + {\alpha ^k}{d^k}
\end{equation}
where $k$ represents the $k$-th iteration. ${\alpha ^k}$ is the step size at the $k$-th iteration, which can be obtained by line search\cite{1999Nonlinear}. ${d^k}$ is the direction of descent, which can be expressed as:
\begin{equation}\label{26}	
	{d^k} =  - \nabla \psi \left( x \right) + {\beta ^k}{d^{k - 1}}
\end{equation}
Where the $\nabla \psi \left( {{x^k}} \right)$ is the gradient of $\psi \left( x \right)$ at ${x^k}$. ${\beta ^k}$ is used to search the new searching direction, defined as:
\begin{equation}\label{27}
	{\beta ^k} = \frac{{{{\left( {\nabla \psi \left( {{x^k}} \right)} \right)}^T}\left( {\nabla \psi \left( {{x^k}} \right) - \nabla \psi \left( {{x^{k - 1}}} \right)} \right)}}{{\nabla \psi {{\left( {{x^{k - 1}}} \right)}^T}\nabla \psi \left( {{x^{k - 1}}} \right)}}
\end{equation}

${\beta ^k}$ ensures that the new search direction is conjugated with the previous one, which is a key feature of the CG algorithm. For the dictionary update problem in Equ.(\ref{24}), the dictionary atoms ${D^k} \in M$ is in manifold space, so the gradient of $\nabla \psi \left( {{D^k}} \right)$ will be a Riemannian gradient. However, we cannot directly combine the $\nabla \psi \left( {{D^k}} \right)$ and $\nabla \psi \left( {{D^k-1}} \right)$, since they belong to different tangent spaces ${T_{{D^k}}}M$ and ${T_{{D^{k - 1}}}}M$. Therefore, we use the Vector $P$ transport method to transport a tangent vector  to a point. The final formula in Equ.(\ref{26}) for the direction update can be expressed as:
\begin{equation}\label{28}
	{d^{{D^k}}} =  - \nabla \psi \left( {{D^k}} \right) + {\beta ^k}{\xi ^{{\alpha ^k}{d^{k - 1}}}}\left( {{d^{k - 1}}} \right)
\end{equation}
Then, the ${\beta ^k}$ can be defined as:
\begin{equation}\label{29}
	{\beta ^k} = \frac{{\left\langle {\nabla \psi \left( {{D^k}} \right),\nabla \psi \left( {{D^k}} \right) - {\xi ^{{\alpha ^k}{d^{k - 1}}}}\left( {\nabla \psi \left( {{D^{k - 1}}} \right)} \right)} \right\rangle }}{{\left\langle {\nabla \psi \left( {{D^{k - 1}}} \right),\nabla \psi \left( {{D^{k - 1}}} \right)} \right\rangle }}
\end{equation}

For $x,y \in {{\rm T}_P}M$, the vector translation map ${\xi ^x}\left( y \right)$ is defined by
\begin{equation}\label{30}
	{\xi ^x}\left( y \right) = \frac{d}{{dt}}{R_p}\left( {x + ty} \right){|_{t = 0}}
\end{equation}	
where ${R_p}\left( . \right) = Ex{p_p}\left( . \right)$ is the exponential map. Based on the above formula, we also need to solve the Riemannian gradient as follows.

According to the formula of the Riemannian gradient, the Riemannian gradient of a tensor $D \in M$ satisfying:
\begin{equation}\label{31}
	\begin{array}{*{20}{c}}
		{{{\left\langle {\nabla \psi \left( D \right),\zeta } \right\rangle }_D} = {{\left\langle {\overline \nabla  \psi \left( D \right),\zeta } \right\rangle }_I},}&\forall
	\end{array}\zeta  \in {T_P}M
\end{equation}
where $\overline \nabla  \psi \left( D \right)$ is the Euclidean gradient of $\psi \left( D \right)$. The Riemannian gradient of i-th dictionary atom is computed by ${D_i}{\overline \nabla  _{{D_i}}}\psi \left( D \right){D_i}$. According to the Equ.(\ref{24}), let ${G_j} = X_j^{ - \left( {1/2} \right)}$ and ${S_j}\left( D \right) = \sum\limits_{i = 1}^n {\alpha _j^i} {D_i}$, then the $\psi \left( D \right)$ can be expressed as:
\begin{equation}\label{32}
	\psi \left( D \right) = \frac{1}{2}\sum\limits_{j = 1}^N {Tr\left( {\log {{\left( {{G_j}{S_j}\left( D \right){G_j}} \right)}^2}} \right)}  + \lambda \sum\limits_{i = 1}^n {Tr\left( D \right)}
\end{equation}

Therefore, the derivative ${\overline \nabla  _{{D_i}}}\psi \left( D \right)$ with respect to atom ${D_i}$ is	
\begin{equation}\label{33}
	\sum\limits_{j = 1}^N {\alpha _j^i} \left( {{G_j}\log \left( {{S_j}\left( B \right)} \right){{\left( {{S_j}\left( B \right)} \right)}^{ - 1}}{G_j}} \right) + \lambda {\rm I}
\end{equation}

Finally, the optimal solution can be solved iteratively by the CG method.}

\bibliographystyle{IEEEtran}
\bibliography{example_paper}

\vspace{11pt}

\vspace{-30pt}
\begin{IEEEbiography}[{\includegraphics[width=1in,height=1.25in,clip,keepaspectratio]{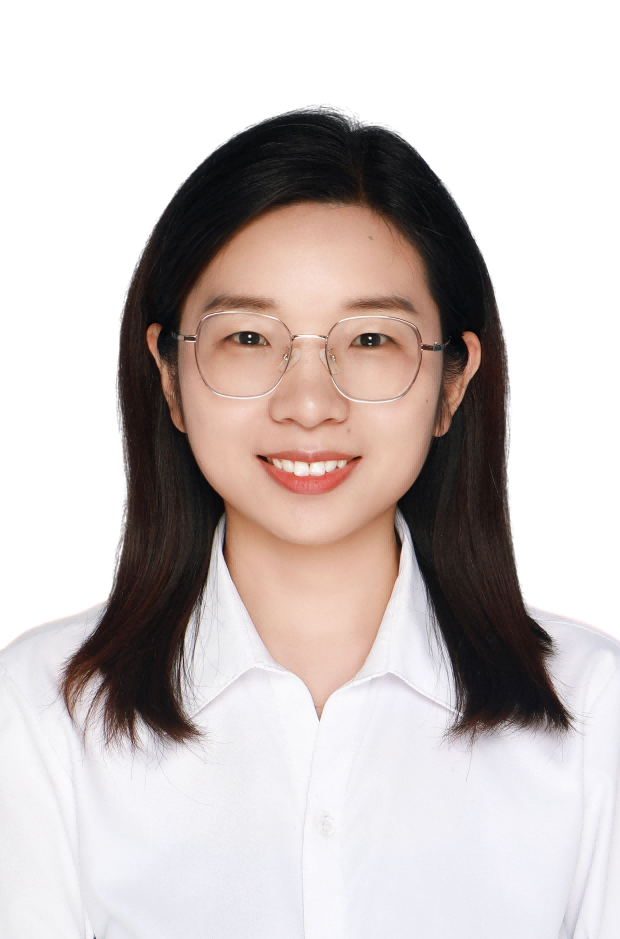}}]{Junfei Shi}
	(Member, IEEE), received the B.S. degree in Computer Science and Technology from Henan Normal University, Henan, China, in 2009, and the Ph.D. degree in computer science and technology from Xidian University, Xi'an, China, in 2016. She is currently a lecturer with the School of Computer Science and Engineering, Xi'an University of Technology, Xi'an.
	
	Her current research interests include polarimetric SAR image classification, semantic model, and computer vision.
\end{IEEEbiography}
\vspace{-30pt}
\begin{IEEEbiography}[{\includegraphics[width=1in,height=1.25in,clip,keepaspectratio]{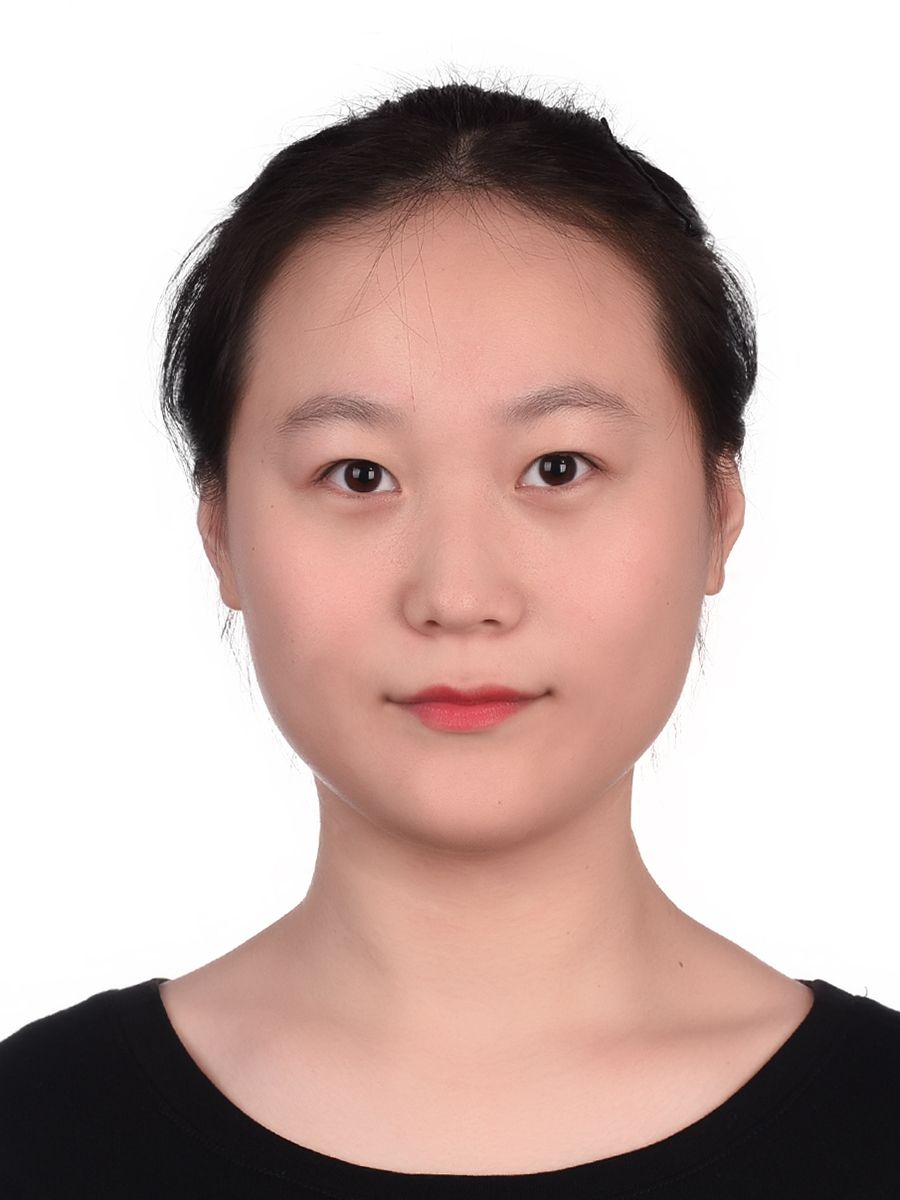}}]{Mengmeng Nie}
	(Graduate Student Member, IEEE), received the B.E. degree in Network Engineering from Xuchang University, Henan, China, in 2022. She is pursuing the M.E. degree in Computer Technology with Xi'an University of Technology, China.
	
	Her main research interests include deep learning and polarimetric synthetic aperture radar (PolSAR) image classification.
\end{IEEEbiography}
\vspace{-30pt}
\begin{IEEEbiography}[{\includegraphics[width=1in,height=1.25in,clip,keepaspectratio]{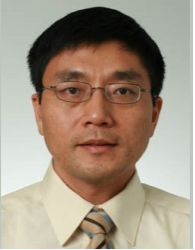}}]{Weisi Lin}
	(Fellow, IEEE) received the bachelor's degree in electronics and the master's degree in digital signal processing from Sun Yat-Sen University, Guangzhou, China, and the Ph.D. degree in computer vision from King's College London, U.K.
	
	He is currently a Professor with the School of Computer Science and Engineering, Nanyang Technological University, Singapore. His research interests include image processing, perceptual modeling, video compression, multimedia communication, and computer vision. He is a fellow of the IET, an Honorary Fellow of the Singapore Institute of Engineering Technologists, and a Chartered Engineer in U.K. He was awarded as a Distinguished Lecturer of the IEEE Circuits and Systems Society from 2016 to 2017. He has served or serves as an Associate Editor for IEEE Transactions on Image Processing, IEEE Transactions on Circuits and Systems for Video Technology, IEEE Transactions on Multimedia, IEEE Signal Processing Letters, and Journal of Visual Communication and Image Representation. He was the Chair of the IEEE MMTC Special Interest Group on Quality of Experience.
\end{IEEEbiography}
\vspace{-30pt}
\begin{IEEEbiography}[{\includegraphics[width=1in,height=1.25in,clip,keepaspectratio]{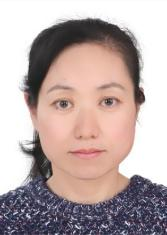}}]{Haiyan Jin}
	(Member, IEEE), is a Professor in Xi'an University of Technology, and she received a Ph.D. degree from Xidian University, China, in 2007. Her research interests include computer vision, image processing and intelligent optimization. She is the winner of Shaanxi Youth Science and Technology Award, and Leader of Shaanxi University Youth Innovation Team.
	
	She has published more than one hundred papers in journals and important international conferences, hosted more than 10 National Natural Science Foundation projects and Shaanxi Province scientific research projects.
\end{IEEEbiography}
\vspace{-30pt}
\begin{IEEEbiography}[{\includegraphics[width=1in,height=1.25in,clip,keepaspectratio]{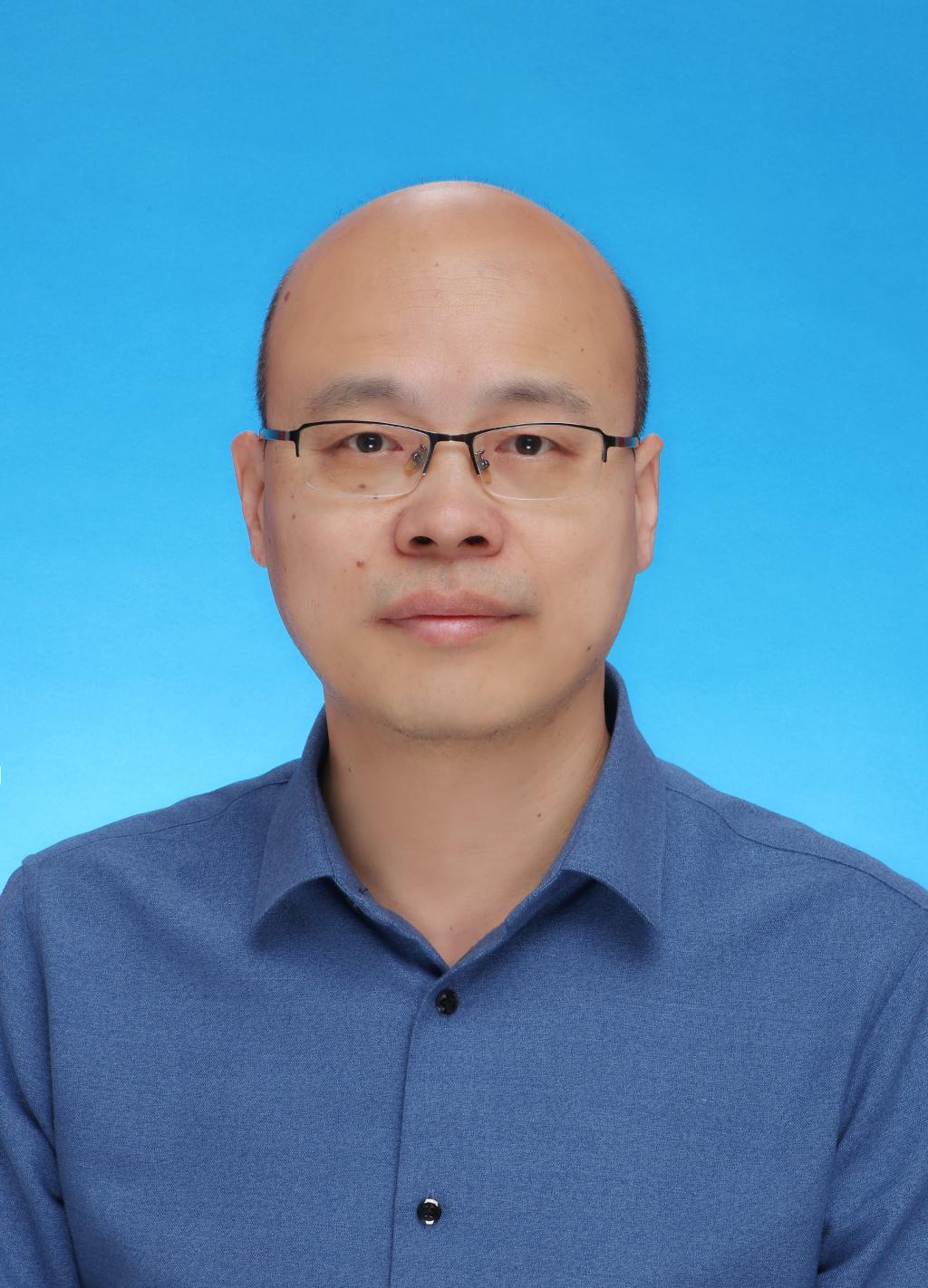}}]{Junhuai Li}
	(Member, IEEE), received the B.S. degree in electrical automation from the Shaanxi Institute of Mechanical Engineering of Xi'an, China, in 1992, the M.S. degree in computer application technology from the Xi'an University of Technology of China, Xi'an, in 1999, and the Ph.D. degree in computer software and theory from the Northwest University of China, Xi'an, in 2002.
	
	He is currently a Professor with the School of Computer Science and Engineering, Xi'an University of Technology, China. His research interests include the Internet of Things technology and network computing.
\end{IEEEbiography}
\vspace{-30pt}
\begin{IEEEbiography}[{\includegraphics[width=1in,height=1.25in,clip,keepaspectratio]{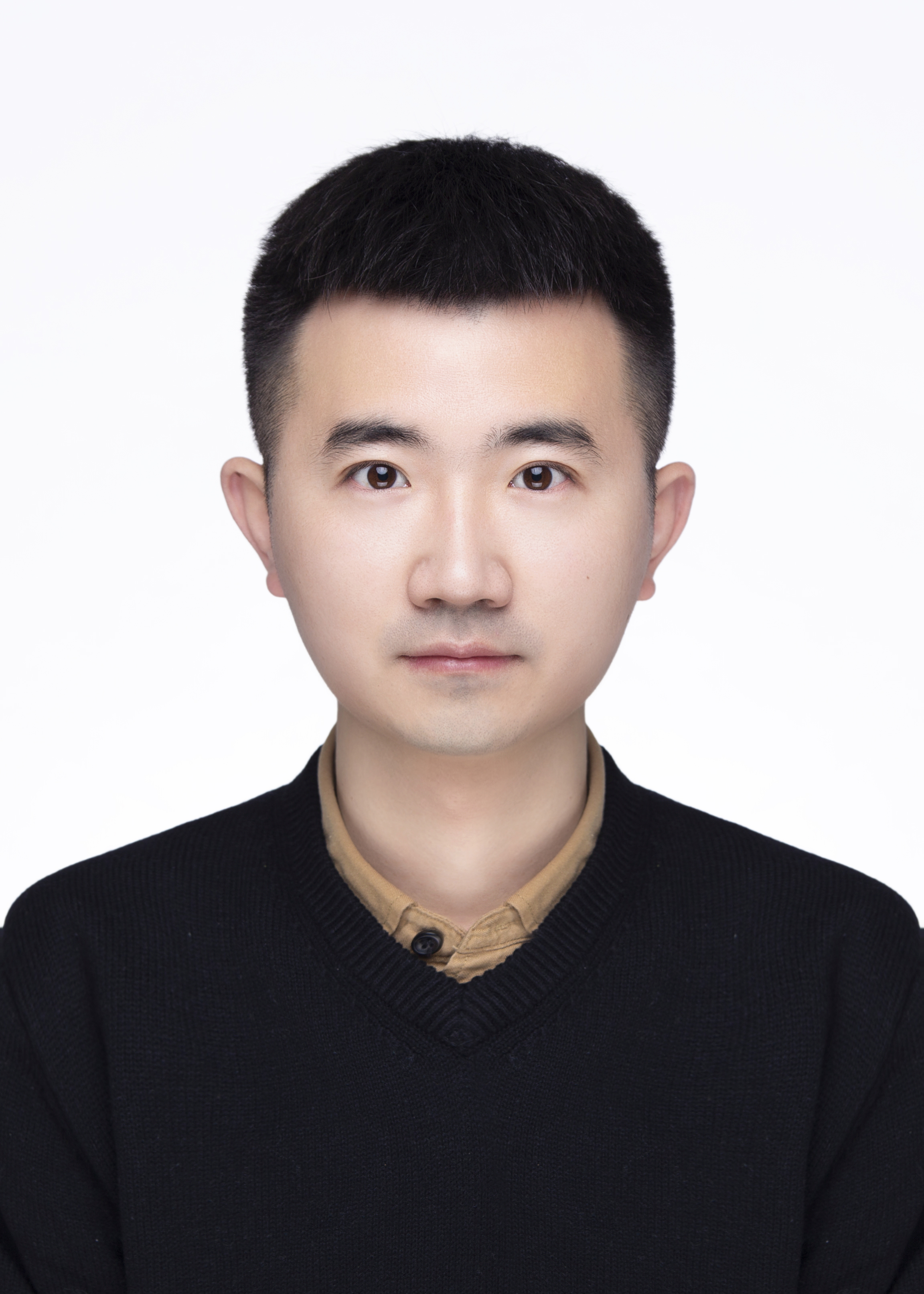}}]{Rui Wang}
	, received the M.S. degree in computer science from Jiangnan University,
Wuxi, China, in 2018, and the Ph.D. degree in  pattern recognition and intelligent system from Jiangnan University,
Wuxi, China, in 2023. He is currently a Lecturer with the School of Artificial
Intelligence and Computer Science, Jiangnan University. His research topics
include Riemannian manifold learning, metric learning, and deep learning. He has published several scientific papers, including TNNLS, TMM, TCSVT, TBD, TCDS, NN, AAAI, IJCAI, ACCV, ICPR etc.
\end{IEEEbiography}
%
\vfill

\end{document}